%% file: main.tex
\documentclass[11pt]{article}
\usepackage{t2t}

\title{T2T Technical Report: Synthesizing GUI Agent Trajectories at Scale from Video Tutorials}

\renewcommand{\RunTitle}{T2T Technical Report: Synthesizing GUI Agent Trajectories at Scale from Video Tutorials}
\renewcommand{\PaperAuthors}{Qingsong Xie\textsuperscript{*,1}, Yanlin Fei\textsuperscript{*,1}, Nazhou Liu\textsuperscript{*,1}, Jiacheng Tang\textsuperscript{1}, Madalina Ciobanu\textsuperscript{1},\\ Navan Preet Singh\textsuperscript{1}, Qingqing Mao\textsuperscript{1,2,\ddag}, Ritankar Das\textsuperscript{1,2}}
\renewcommand{\PaperAffil}{\textsuperscript{1}Prentis AI, San Francisco, CA\quad\textsuperscript{2}Titan Holdings, San Francisco, CA}
\renewcommand{\TitleFootnote}{%
  \textsuperscript{*}Equal contribution.\quad
  \ifshowlead\textsuperscript{\dag}Project lead.\quad\fi
  \textsuperscript{\ddag}Corresponding author.}

\showdatatrue
\renewcommand{\DataLine}{{\IconData}~Data: \href{https://huggingface.co/datasets/PrentisAI/t2t-v1}{PrentisAI/t2t-v1}}
\renewcommand{\MailLine}{{\IconMail}~qmao@prentis.ai}

\usepackage{parskip}

\begin{document}
\maketitle

\input{sections/00_abstract}

\input{sections/01_introduction}
\input{sections/02_pipeline}

\input{sections/05_training}

\input{sections/07_results}

\input{sections/09_conclusion}

\bibliographystyle{plainnat}
\bibliography{refs}

\clearpage
\appendix
{\centering\Large\bfseries\color{t2taccent}Appendix\par}
\vspace{0.5em}{\color{black}\hrule height0.6pt}\vspace{1.2em}
\suppressfloats[t]   
\input{sections/99_appendix}

\end{document}

%% file: sections/00_abstract.tex
\begin{abstract}
\noindent
Training generalist GUI agents requires large volumes of interaction trajectories comprising screenshot sequences paired with grounded UI actions, yet collecting such data through human demonstration is expensive and difficult to scale. Although large-scale collections of public software tutorial videos contain rich operational demonstrations, their continuous pixel streams lack discrete action boundaries, semantic intents, and precise spatial coordinates, making them difficult to use directly for model training. To address this, we present \textbf{T2T} \textbf{(Tutorial-to-Trajectory)}, a fully automated pipeline for synthesizing GUI agent trajectories at scale. T2T spans desktop, web, and mobile platforms to convert unannotated video tutorials into structured, executable interaction trajectories. To guarantee trajectory quality, T2T combines active segment filtering, keyframe identification via localized UI change, shortcut-free observation frame selection, presence-aware grounding, and two-stage action-space normalization. These refined trajectories are then converted into multi-task training views spanning navigation, planning, forward dynamics, and grounding. We fine-tune Qwen3-VL-8B on a subset of the more than one million human-annotation-free training samples generated by this pipeline. Evaluation results show that the model achieves consistent gains across multiple benchmarks covering desktop, web, and mobile environments, including OSWorld-G, ScreenSpot-Pro, AndroidControl, and AgentNetBench, indicating improved core capabilities in grounding, step-level instruction execution, and trajectory-level decision-making. Notably, all the aforementioned gains are achieved without introducing any in-distribution training data from target domains. By learning exclusively from the general interaction trajectories extracted from public videos via the T2T pipeline, the model exhibits cross-application and cross-task generalization capabilities.
\end{abstract}

%% file: sections/01_introduction.tex
\section{Introduction}

Large language models and vision-language models have demonstrated remarkable capabilities in understanding natural language, generating code, and reasoning over visual content~\citep{openai2024gpt4o, bai2025qwen25vl, chen2024internvl25}, increasingly serving as general-purpose foundations for a wide range of intelligent applications. A natural next step is to translate these capabilities into concrete task execution in the digital world. Unlike web APIs or command-line interfaces that expose structured endpoints, the vast majority of software, from productivity suites and creative tools to enterprise systems and mobile applications, is designed primarily for human interaction through graphical user interfaces. This makes GUI the de facto universal interface to the digital world, but also one that demands perceptual and motor skills traditionally reserved for human users: recognizing UI elements from pixels, understanding spatial layouts, tracking state changes across screens, and executing precise sequences of clicks, keystrokes, and gestures~\citep{cheng2024seeclick, gao2025uground, xie2026groundgui}. GUI agents, autonomous systems built on top of LLMs and VLMs that perceive screen content and perform such actions on behalf of users, bridge this gap by operating directly at the interface level, automating tasks across arbitrary applications without requiring dedicated APIs or developer-side instrumentation. A substantial body of work has advanced this direction in recent years~\citep{wang2025guisurvey, zhou2024guisurvey}, producing increasingly capable agents across desktop, mobile, and web platforms~\citep{lin2025showui, xu2025aguvis, uitars2_2025, liu2025scalecua, xu2026mobilerl, li2026uir1, xiao2024autoglm}.

Despite rapid progress in model architecture and training recipes, the availability of high-quality trajectory data remains a central bottleneck. Training a capable GUI agent requires large collections of expert demonstrations: sequences of screenshots paired with precise actions that together capture realistic, multi-step interactions with diverse applications and interfaces. Curating such data through human annotation is inherently expensive and slow, and the cost grows sharply with task complexity and horizon length.

A natural alternative source of such data is video tutorials. Platforms such as YouTube host millions of screen recordings in which human experts demonstrate complex software workflows step by step, spanning a broad range of applications across desktop, mobile, and web platforms. These videos are inherently long-horizon, reflect realistic human operating patterns, and are available at a scale that manual annotation cannot easily match.

Several recent efforts have recognized the potential of web-sourced tutorials and videos, and explored converting them into agent trajectories~\citep{lu2025videoagenttrek, xiong2026video2gui, zhang2025tongui, song2025watchlearn, xie2026guide}. However, bridging the gap from raw video to structured, training-ready trajectories presents several technical challenges: \textbf{(1) Computational and Contextual Limits.} Feeding unsegmented raw videos into vision-language models (VLMs) is computationally prohibitive and practically infeasible due to strict API limits and context windows. Pruning the vast majority of frames that contain non-interactive content (e.g., introductions, static slides, or idle pauses) and identifying informative frames are essential for efficient and effective video processing. \textbf{(2) Visual Target Leakage and Shortcut Learning.} Naively sampled screenshots often suffer from severe visual shortcuts. High-production tutorials frequently overlay colored boxes or arrows, training models to locate annotations rather than comprehend UI pixels. Similarly, taking screenshots at the exact action timestamp leaves the cursor directly hovering over the target, encouraging models to learn a trivial ``read-the-cursor'' shortcut. \textbf{(3) Coordinate Hallucination under Target Absence.} Standard grounding models may instead rely on layout priors (e.g., close buttons in corners) and hallucinate coordinates even when the intended target element is temporarily absent or occluded, injecting noisy coordinates into the training distribution. \textbf{(4) Action Space Heterogeneity and Misalignment.} Speech narration and screen activity are temporally asynchronous. Furthermore, raw video extraction produces a heterogeneous vocabulary of non-standardized action names, which must be mapped into unified, platform-conditioned tool-call schemas.

In this report, we present \textbf{T2T} (Tutorial-to-Trajectory), a fully automated pipeline designed to overcome these challenges. Starting from raw public videos, the pipeline proceeds through three stages:
\textbf{(1) Video Filtering and Preprocessing.} We prune non-interactive video segments using cursor-based sparse probing as a user activity proxy, and enforce strict visual annotation filtering to eliminate superimposed highlights, arrows, and bounding boxes that cause label leakage.
\textbf{(2) Trajectory Extraction and Action Spatial Grounding.} Instead of relying on noisy global frame differences or asynchronous audio transcripts, we pinpoint discrete action timestamps by scoring localized UI changes (explicitly discounting full-screen turnovers) and extracting settled before/after frame pairs. A vision-language model then derives task context, sub-step plans, and target referring expressions. To prevent the model from simply memorizing pointer positions, we perform a backward search for pre-action screenshots where the cursor is away from the target element. A three-tier presence-aware grounding cascade then localizes target elements, explicitly discarding unverified coordinates to prevent label noise while repurposing rejected steps as negative abstention samples.
\textbf{(3) Action Normalization and Multi-task Conversion.} We normalize heterogeneous extractor action vocabularies through a two-stage normalization scheme into unified, platform-conditioned tool-call schemas (computer use, browser use, and mobile use). Finally, trajectories are rendered into complementary multi-task training views (navigation, planning, forward dynamics, and grounding) to maximize data efficiency.

Applied to a large-scale collection of public tutorial videos, T2T yields over one million training samples without human annotation. Supervised fine-tuning of Qwen3-VL-8B on a ${\sim}350$K mixture drawn from this corpus, without utilizing any target-domain or benchmark training data, yields the fine-tuned model T2T-8B, which improves over the base model across multiple benchmarks: raising GUI grounding accuracy on OSWorld-G and ScreenSpot-Pro, boosting step success rate on AndroidControl, and outperforming the base model on AgentNetBench. These results show that a rigorous pipeline can enable raw video tutorials to serve as a scalable and effective data engine for training generalist GUI agents.

%% file: sections/02_pipeline.tex
\section{The T2T Pipeline}
\label{sec:pipeline}

\subsection{Overview}

\begin{figure}[htbp]
    \centering
    \includegraphics[width=0.98\textwidth]{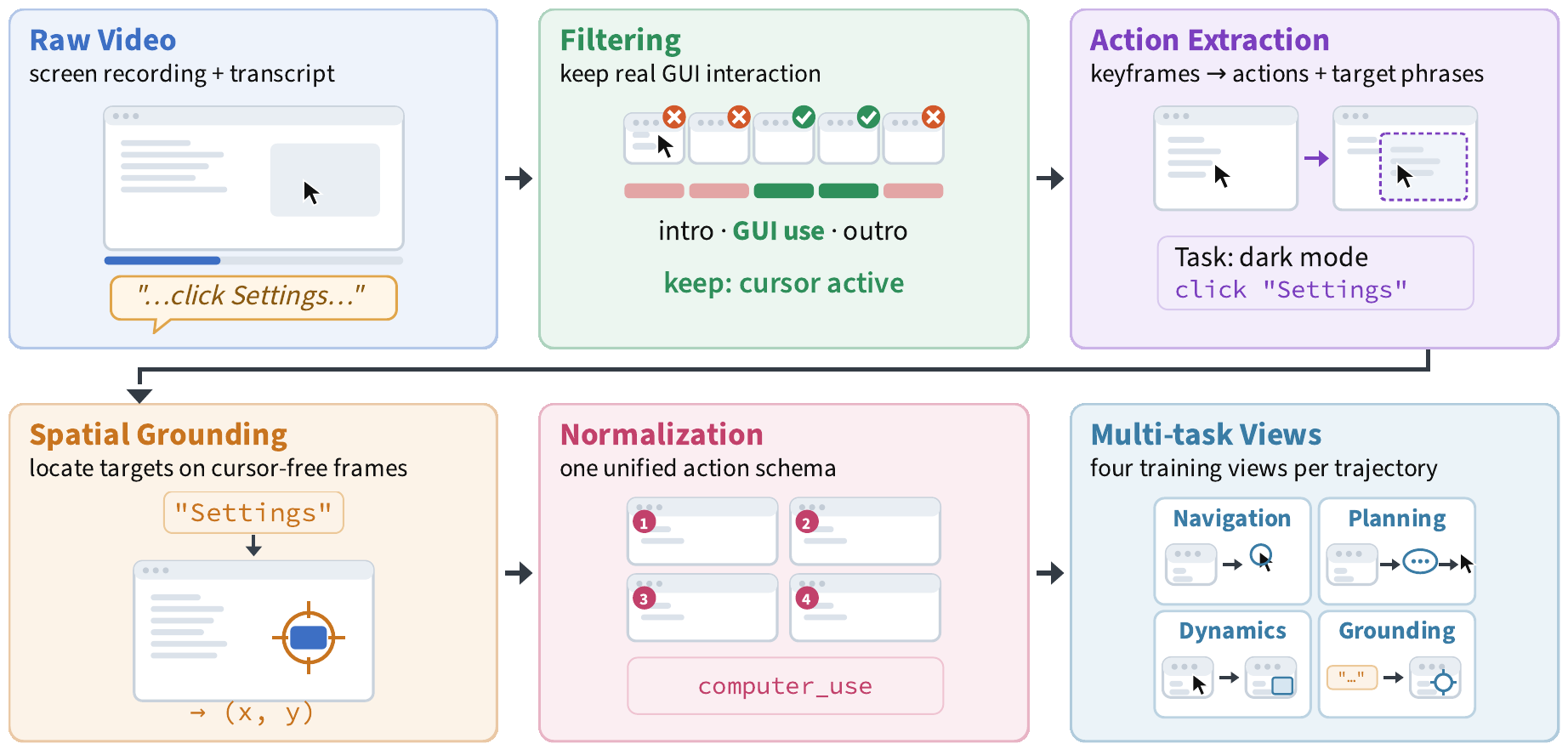}
    \caption{\textbf{Overview of the T2T (Tutorial-to-Trajectory) pipeline.} T2T converts GUI tutorial videos into structured interaction trajectories through video filtering, action extraction, spatial grounding, and action normalization. The resulting trajectories are transformed into four training views: navigation, planning, forward dynamics, and grounding.}
    \label{fig:pipeline}
\end{figure}

\Cref{fig:pipeline} summarizes the pipeline. Given a screen recording and its transcript, T2T must answer three questions that the raw video does not state: \emph{when} the user is interacting with the interface at all, \emph{that} an action occurred and of what type, and \emph{where} on the screen it was aimed. The first is answered during video filtering and segmentation (\Cref{sec:filtering}); the second and third during trajectory extraction and grounding (\Cref{sec:extraction}), which also groups the recovered actions into tasks. A final stage (\Cref{sec:actionspace}) expresses the recovered actions in a single schema and renders them into training samples. Because the three inferences fail independently and at different rates for different action types, every stage records how each decision was reached, not only its outcome; this provenance (\Cref{app:provenance}) is what allows the corpus to be filtered and audited after the fact.

\subsection{From Video to Informative Segments: Video Preprocessing and Filtering}
\label{sec:filtering}

\begin{figure}[htbp]
    \centering
    \includegraphics[width=0.98\textwidth]{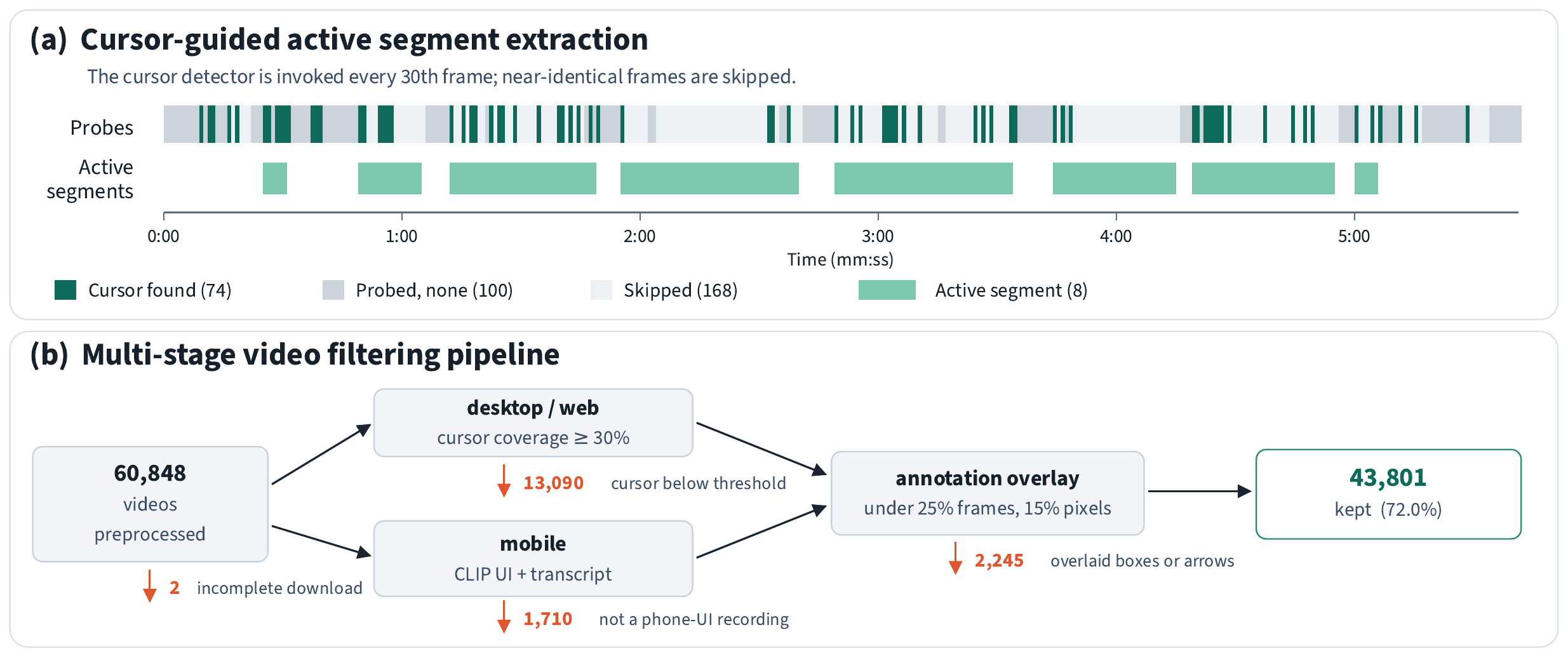}
    \caption{\textbf{Video filtering and preprocessing.} (a) Cursor-guided extraction of active interaction segments using sparse, adaptive cursor detection. (b) Multi-stage filtering of desktop, web, and mobile tutorials based on GUI activity, visual and semantic cues, and annotation overlays.}
    \label{fig:filtering}
\end{figure}

\begin{figure}[!t]
    \centering
    \includegraphics[width=0.9\textwidth]{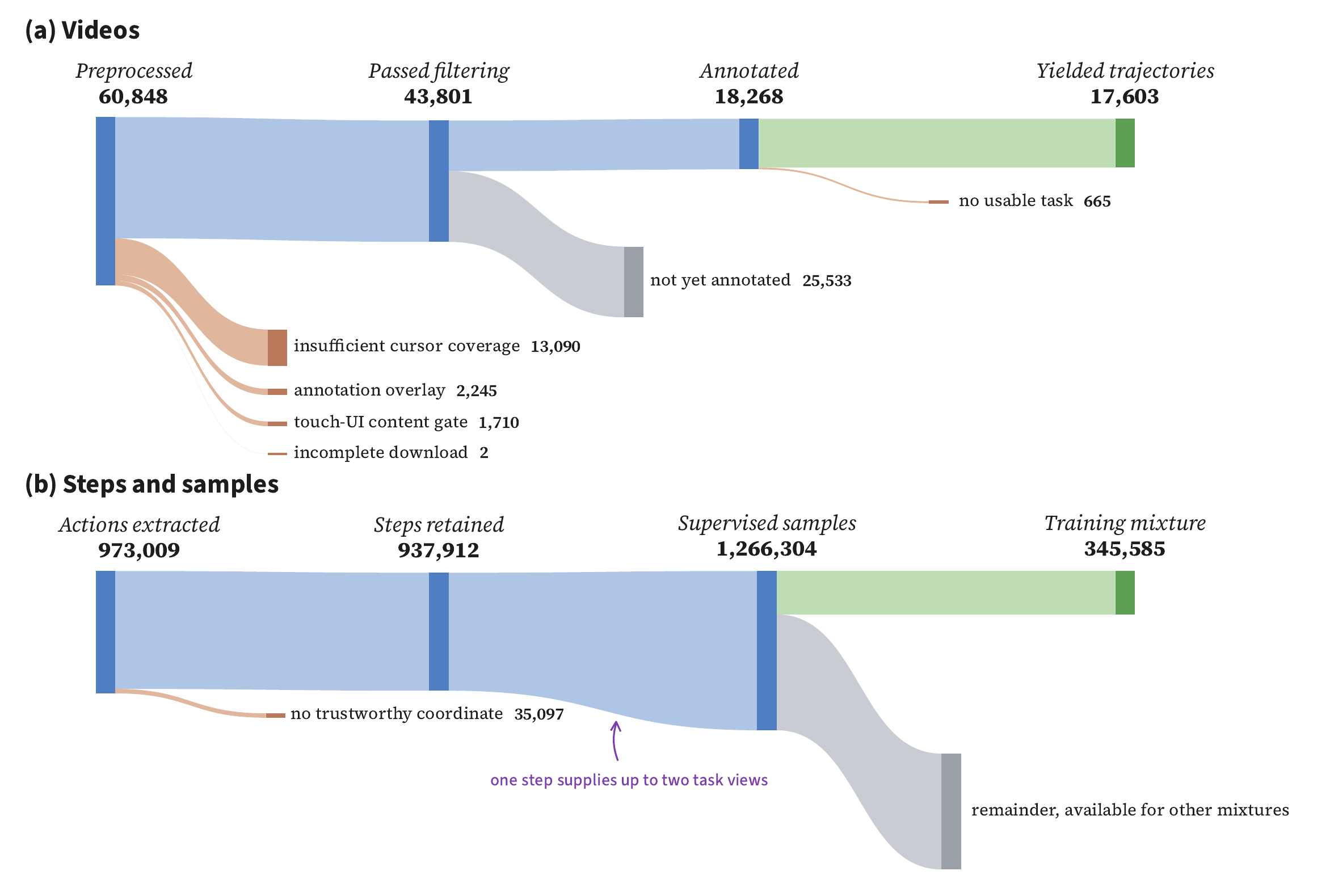}
    \caption{\textbf{Data flow through the T2T pipeline.} (a) Video filtering and trajectory extraction. Some videos are processed at more than one resolution, so the nodes count preprocessing runs rather than unique videos: the 17{,}603 runs that yield trajectories correspond to the 16{,}974 unique source videos of \Cref{sec:corpus}. (b) Conversion of extracted actions into retained steps, supervised samples, and the final training mixture. Supervised samples are counted over the navigation and grounding views, from which the training mixture is drawn; with the planning and forward-dynamics views the corpus yields 2{,}709{,}896 samples (\Cref{sec:actionspace}). Brown branches indicate rejected data, while gray branches indicate data held in reserve.}
    \label{fig:data_flow}
\end{figure}

Web-sourced videos provide a large pool of potential training data, but their actual utility varies widely. Many clips feature no human-GUI interaction at all, including talking-head lectures, promotional teasers, and static slides. Even genuine screen tutorials are frequently padded with idle pauses, title cards, and off-topic discussion. Feeding raw footage directly into a vision-language model is computationally wasteful, clogging the model's context with tens of thousands of uninformative frames. Worse, non-interactive clips dilute training quality by breaking the clear mapping between user actions, such as clicks or typing, and visual interface changes. This stage addresses both challenges by filtering out irrelevant videos and
extracting \emph{informative segments}---spans of continuous human--GUI
interaction---from the videos that are retained. \Cref{fig:filtering}
illustrates both steps, and \Cref{fig:data_flow} traces the resulting data
volumes through the full pipeline.

\subsubsection{Cursor-based Segmentation and Filtering}
In desktop and web tutorial videos, cursor movement serves as a primary indicator for human-GUI interaction. We track cursor activities to pinpoint active segments from raw video footage.

\textbf{(1) Detecting Cursor Presence.} Cursor glyphs typically span only a few dozen pixels on high-resolution displays, vary widely in shape from standard arrows and I-beams to hand pointers and custom icons, and often blend into complex UI backgrounds. Therefore, we deploy a dedicated cursor detector operating at high input resolutions to improve cursor detection accuracy. Given a single frame from a screen recording, the detector outputs the bounding box of the cursor pointer alongside a confidence score, or yields no detection when the cursor is absent. We apply a permissive confidence threshold to prioritize high recall, as a false positive merely extends an informative segment slightly, whereas a false negative prematurely truncates an ongoing human-GUI interaction.

User interactions usually span several seconds, making consecutive video frames nearly redundant as evidence of activity. Running cursor detection on every frame is therefore computationally wasteful. We replace full framerate inference with a sparse sampling schedule of roughly one frame per second. Furthermore, we make this probing adaptive by skipping the detector whenever the current frame matches the visual state of the last frame where a cursor was found. In these cases, the skipped frames are automatically assumed to contain a cursor. As a static screen provides no new interaction signals, this mechanism eliminates unnecessary computation during idle pauses.

\textbf{(2) Extracting Informative Segments.} The cursor detector outputs a series of discrete frames where a pointer is present, but downstream processing requires continuous segments. Defining segment boundaries merely by adjacent positive frames (those with detected cursors), is insufficient, as the pointer often vanishes intermittently during active usage, such as when occluded by dropdown menus, parked off screen, or kept still during reading pauses. To accommodate this intermittent visibility, we tolerate short non-detection gaps using a defined grace period. A segment opens at the first detected frame, remains active across transient non-detection gaps, and closes only after this specified period elapses without any new detection. Finally, we prune segments falling below a minimum duration threshold to eliminate accidental mouse jitters and video editing artifacts.

\textbf{(3) Filtering Videos Based on GUI Activity Duration.} To decide whether a video should be retained, we evaluate its cumulative informative segment coverage rather than its raw frame-level detection rate. Specifically, a video is kept if its extracted informative segments account for a sufficient fraction of its total duration. This coverage-based metric is essential because our segment extraction already bridges short pauses when the cursor is temporarily hidden or stationary. Scoring videos purely on raw cursor hits would falsely reject high-quality tutorials that contain minutes of continuous workflow. Finally, the temporal boundaries of these informative segments are passed downstream so that the subsequent trajectory extraction stage (\Cref{sec:extraction}) processes only valid interaction intervals. The operating points of all filters are listed in \Cref{tab:filterparams}.

\subsubsection{Visual and Semantic Filtering for Mobile Videos}

Mobile tutorial videos lack cursor signals because touch input replaces mouse pointers. To filter mobile recordings effectively, we enforce a dual-validation pipeline where a video must satisfy both visual and semantic criteria. First, the visual filter uses a zero-shot image classifier on uniformly sampled frames to ensure that at least half of the sampled frames actually display a mobile UI. Second, the semantic filter analyzes the video transcript to confirm that imperative instructional phrases occur with sufficient frequency (captions and phrase list in \Cref{app:p-clip}). Enforcing both conditions eliminates two common failure modes, namely videos that show mobile screens while lacking actionable guidance, and videos that provide tutorial narration while mostly featuring presenter footage.

\subsubsection{Visual Annotation Filtering}

Beyond the above approaches, we apply an additional filter to prevent label leakage. High-production tutorials frequently superimpose bright rectangles, arrows, or highlights over target UI elements. These visual overlays effectively embed ground-truth answers directly into image pixels. As a result, a model may learn a superficial shortcut by tracking the annotations rather than understanding the interface, causing it to fail on clean screenshots during inference. To address this, we sample video frames and reject candidates using two complementary rules (thresholds in \Cref{tab:filterparams}). First, we detect structured bounding-box outlines across a significant portion of frames. Second, we measure the total pixel area occupied by characteristic annotation colors. This dual mechanism is essential because the first rule identifies crisp box boundaries, while the second catches diffuse highlights and freehand markings that evade box detection.

\subsection{From Video Segments to Grounded Trajectory: Trajectory Extraction and Action Spatial Grounding}
\label{sec:extraction}

\begin{figure}[!t]
    \centering
    \includegraphics[width=0.96\textwidth]{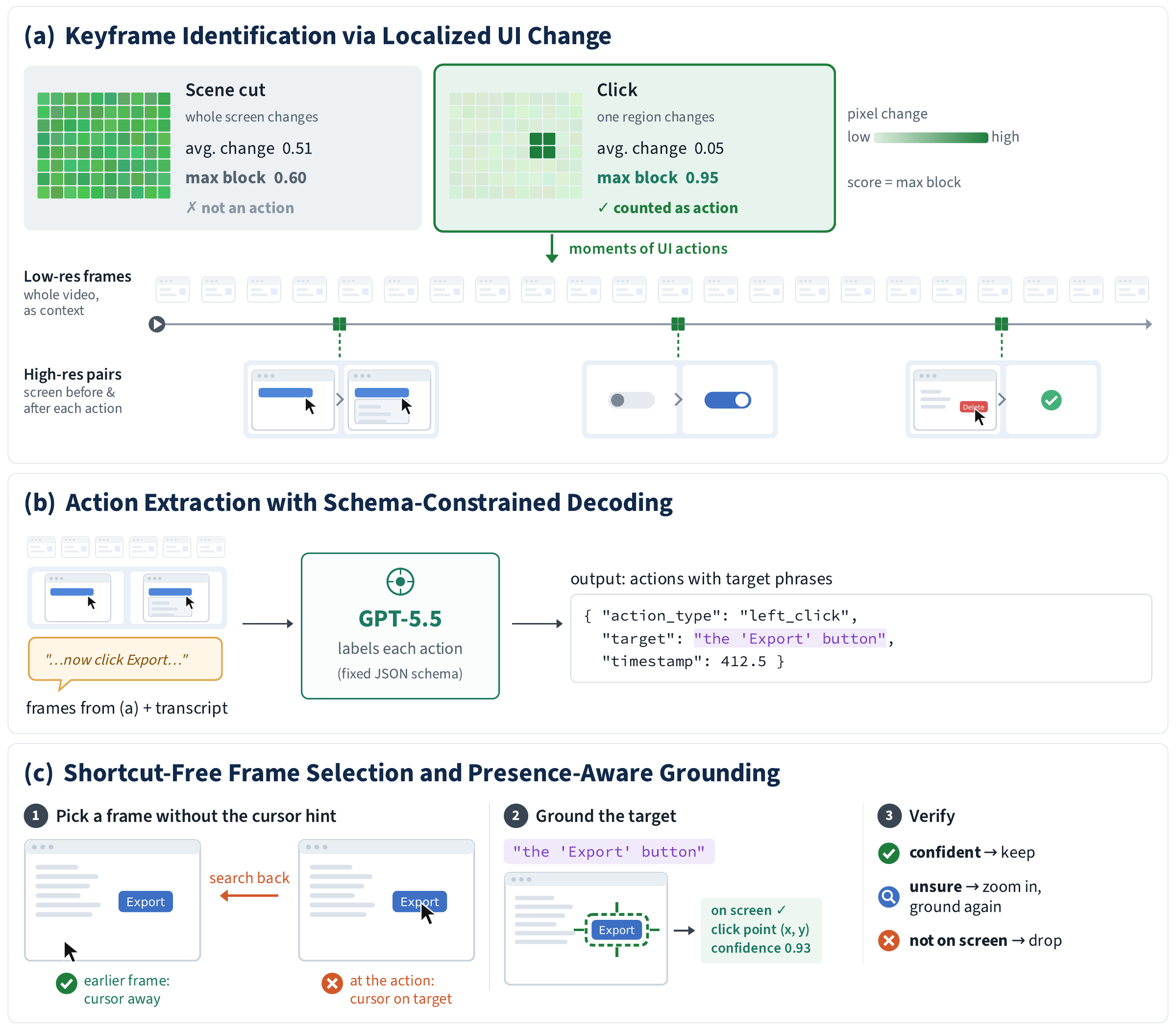}
    \caption{\textbf{Trajectory extraction and spatial grounding.}
(a) Keyframe identification based on localized UI changes, combining high-resolution before–after pairs with low-resolution contextual frames. (b) Schema-constrained extraction of tasks and actions with semantic target descriptions. (c) Shortcut-free observation selection and presence-aware spatial grounding with confidence-based verification.}
    \label{fig:trajectory_extraction}
\end{figure}

Having isolated informative human-GUI interaction segments from raw videos, we now convert these visual clips into structured agent trajectories. On their own, unannotated video segments lack three essential elements required for a trajectory: pre-action visual states describing what the interface looks like before an interaction, textual action intents explaining why the action is performed, and spatial interaction coordinates specifying where the interaction occurs. To bridge this gap, this stage transforms the filtered segments into a discrete execution trajectory, mapping each step to a pre-action interface state $S_t$, a textual action description $A_t$, and grounded spatial coordinates $C_t$.
\Cref{fig:trajectory_extraction} illustrates the main steps.

\subsubsection{Keyframe Identification via Localized UI Change}

To convert continuous video segments into discrete agent trajectories, we identify keyframes that capture the precise visual states immediately before and after each user interaction. The pre-action frame $S^{\text{before}}$ serves as the visual context for agent decision-making, while the post-action frame $S^{\text{after}}$ reveals the state transition required to infer action intent and verify execution. Pinpointing these transition moments is challenging because direct fixed-interval sampling blindly designates frames extracted at uniform temporal increments as trajectory steps. By ignoring natural action timing, this approach frequently captures mid-animation transient frames or skips brief interactions altogether, creating temporal misalignment between visual states and actual user actions. On the other hand, relying on pixel differences between consecutive full frames to detect action occurrences is unreliable. Global background movements, such as page scrolling or video cuts, generate massive frame-wide variance that easily drowns out the subtle localized changes produced by genuine UI interactions. 

We resolve this through a two-stage strategy. First, we identify candidate change moments, defined as estimated timestamps in the video timeline where a GUI interaction is likely triggered. Rather than scanning every raw frame, we coarsely sample video frames and compute initial transition scores by partitioning each frame into a spatial grid, evaluating maximum localized pixel change while discounting full-screen turnovers. To refine these initial scores, we incorporate two multimodal priors. Candidate scores are multiplied by a boost factor if they fall within a temporal window surrounding transcript action verbs such as click or tap. Furthermore, an object detector locates the cursor bounding box, scaling up scores for candidates where UI changes occur directly adjacent to the pointer while discounting distant changes. All candidate moments are then re-ranked, and the top-scoring moments across the video timeline are selected for keyframe extraction. Second, for each selected top-scoring moment, we dynamically scan neighboring frames in the native video stream to locate stable boundaries where temporal pixel variations drop below a threshold, capturing clean $S^{\text{before}}$ and $S^{\text{after}}$ states once interface animations subside. 

These sparse high-resolution keyframe pairs are interleaved with a dense stream of low-resolution frames sampled at fixed intervals across the entire video recording. While the high-resolution pairs supply fine-grained details for localized GUI interaction, the low-resolution stream captures comprehensive background context for the entire task. This interleaved sequence is subsequently passed to the downstream action extraction module.

\subsubsection{Action Extraction and Schema-constrained Decoding}
\label{subsubsection: action extraction}
Given the interleaved sequence of sparse keyframe pairs and dense background context, action inference is powered by GPT-5.5 (prompts in \Cref{app:p-extract}). When an input frame sequence exceeds the context length limit, it is processed through sequential temporal windows. Compact structured summaries from previous windows are passed forward, ensuring that tasks spanning window boundaries maintain logical continuity without losing state history.

To prevent cross-domain action hallucinations, where models mistakenly output invalid actions from other environments such as mobile swipe gestures on web interfaces or platform-incompatible shortcuts, model decoding is constrained by a platform-specific schema that limits outputs strictly to valid native operations. The model generates structured action parameters, including action type, frame timestamps, typed text strings, key combinations, and scroll vectors, together with a referring expression for the target element. To eliminate spatial ambiguity during downstream grounding, each referring expression must identify the smallest actionable control using its literal interface text, such as the Export button, rather than referencing broad structural containers like toolbars or sidebars.

Individual actions are hierarchically grouped into sub-tasks, each labeled with a high-level user instruction and a concise execution plan. Because every action carries a specific task identifier, a single video containing multiple independent workflows is cleanly divided into separate trajectories.

\subsubsection{Shortcut-Free Observation Frame Selection}
\label{subsubsection: observation frame}

Each extracted action must be paired with an appropriate pre-action observation frame $O_t$ that reflects what an agent would observe before initiating an interaction. While the keyframe extraction stage captures $S^{\text{before}}$ immediately preceding a state transition, directly using $S^{\text{before}}$ as $O_t$ introduces two critical failure modes that may impair trajectory dataset quality.

First, interaction-induced element disappearance occurs when the target UI control vanishes during the interaction itself. In modern user interfaces, elements such as dropdown menu items, context menus, hover tooltips, and confirmation popups immediately collapse or close upon being clicked. If the $S^{\text{before}}$ frame is captured at the precise moment of action trigger or during a slight temporal overlap, the target element may already be absent from the display, leaving downstream agents without a visible interaction target in their observation frame.

Second, cursor position leakage leads to shortcut learning, where the mouse pointer corrupts the visual observation. Because $S^{\text{before}}$ is captured immediately prior to action execution, the cursor is often resting directly on top of the target UI element. When downstream models are trained on such frames, they may quickly learn a spurious shortcut by tracking the cursor location to identify the target, rather than parsing the visual interface and grounding textual referring expressions.

To resolve both failure modes, we search backward from the action timestamp for the latest video frame that satisfies three strict criteria. The candidate frame must maintain structural consistency with the action frame to ensure the overall interface layout has not shifted. The target UI element must remain fully visible and unoccluded. Finally, the cursor must be displaced away from the target element, ensuring that the pointer does not overlap with the target bounding box. When no frame satisfies all three criteria, the default $S^{\text{before}}$ frame is retained as $O_t$ and the step is flagged for quality control.

\subsubsection{Presence-Aware Spatial Grounding}
\label{sec:grounding}

Given the pre-action observation frame $O_t$ selected in the preceding step and the textual referring expression of the target UI control extracted from the action pipeline, mapping these inputs to spatial coordinates $C_t$ is formulated as single-image visual grounding. Standard grounders often output coordinates even when the target element is absent by relying on layout priors, such as assuming a close button resides near the top-right corner, thereby injecting confident errors into the training distribution. We instead require explicit target-presence verification, allowing the system to decline prediction when an element is missing.

This mechanism is operationalized as a three-tier adaptive cascade (prompts for each tier in \Cref{app:p-ground}). First, a fast pass feeds the full observation frame $O_t$ and the target textual expression into Qwen3.7-Plus without chain-of-thought reasoning. This pass employs a high-recall prompt that encourages candidate generation whenever a plausible control is observed, rapidly producing a structured JSON response containing a presence flag, coordinate points, and a confidence score. When this pass yields a low-confidence prediction with a score below 0.6, a localized visual patch centered at the predicted coordinate is cropped from $O_t$ and evaluated by a locally hosted Qwen3.6-27B model using a focused binary presence prompt. Predictions that are initially declined, as well as candidates explicitly rejected by the crop verification model, are escalated to a reasoning-enabled pass. Here, Qwen3.7-Plus re-evaluates the full observation frame $O_t$ under a conservative prompt that explicitly instructs the model to assume a high likelihood of target absence. Guided by this strict prompt together with chain-of-thought reasoning, the model delivers a final presence and coordinate determination.

Verification fails when the escalated reasoning pass also declines presence, whether it was reached because the fast pass declined or because the cropped region did not contain the target. Some of these failures stem from source video characteristics, such as a transient dropdown menu or dialog already closing, a target element scrolling out of view, or a referring expression naming a structural container rather than an actionable control. Other failures arise from technical issues, such as unparseable model responses or transport errors, which are separated by a downstream filter before discarded steps are put to further use. Failure handling then depends strictly on action semantics. Pointing actions require precise coordinates and are removed from the trajectory, text entry actions degrade gracefully to ungrounded steps, and scroll operations are retained because their coordinates specify broad screen regions rather than localized controls. Finally, non-spatial operations such as hotkey combinations and wait commands bypass spatial grounding entirely.

\begin{figure*}[!t]
\centering
\includegraphics[width=\textwidth]{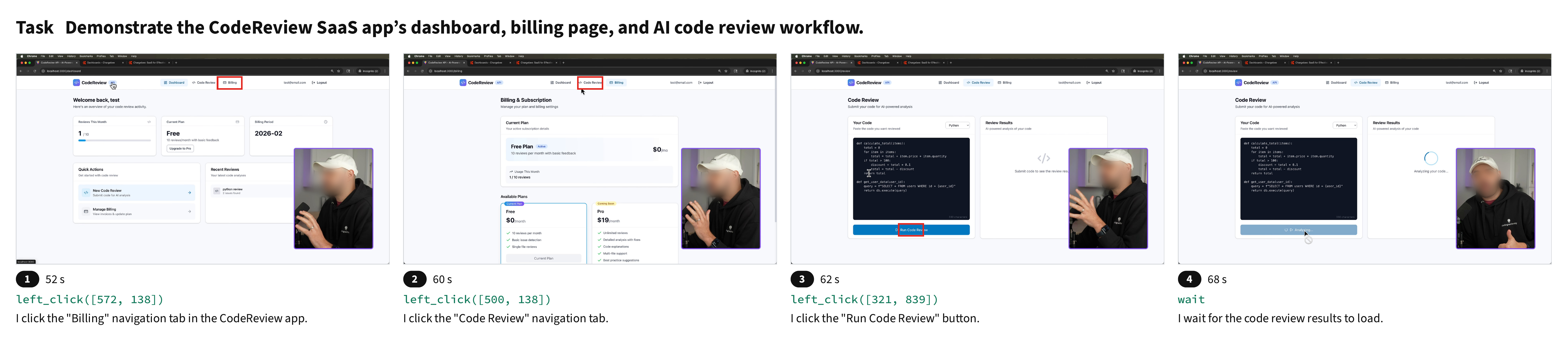}
\caption{\textbf{Example of an extracted interaction trajectory. } Each step includes a pre-action observation, source-video timestamp, normalized tool call, and textual action description. Red markers indicate grounded interaction coordinates.}
\label{fig:trajectory}
\end{figure*}

\subsubsection{Trajectory Assembly}

Extracted steps are grouped by sub-task into ordered trajectories, discarding single-action tasks that offer no sequential learning signal. Action timestamps are aligned directly with the visual frames processed during extraction, enabling deterministic frame retrieval from the video stream for every step. Spatial coordinates are projected into a normalized coordinate space, decoupling trajectory representations from both the original capture resolution and any downstream resizing. When an intermediate action is removed due to failed spatial verification, its textual description is preserved within the interaction history of subsequent steps, maintaining chronological task continuity while preventing models from training on unverified coordinates. \Cref{fig:trajectory} illustrates an assembled trajectory as stored in the final corpus; complete stored samples are given in \Cref{app:sample}.

\subsection{From Trajectories to Training-ready Tasks}
\label{sec:actionspace}

\subsubsection{Action-Space Normalization}

While schema-constrained decoding during action extraction guarantees that extracted outputs conform to syntactical types, it cannot prevent semantically degenerate arguments or reconcile heterogeneous vocabularies across schema revisions and pipeline iterations. Decoupling raw action extraction from downstream training specifications allows action schemas and tool definitions to evolve without re-executing computationally expensive visual inference across the video corpus.

To achieve this, normalization proceeds through a two-stage pipeline. First, raw outputs from the extraction stage are preserved verbatim in trajectory logs as detailed in \Cref{tab:nativefull}, ensuring complete trajectory auditability and supporting future schema revisions without re-annotating the underlying dataset. Second, each step is mapped to a JSON tool invocation under one of three platform-conditioned schema definitions, namely \texttt{computer\_use}, \texttt{browser\_use}, and \texttt{mobile\_use}, as described in \Cref{tab:mobileuse} and \Cref{tab:computeruse and browseruse}. This second stage canonicalizes variant synonyms into unified primitives (full mapping in \Cref{tab:mappingfull}), filters out degenerate operations such as zero-distance drags or zero-magnitude scrolls, and enforces platform-conditioned routing when the same high-level intent requires different platform-specific executions, such as resolving a back action to a native system gesture on mobile interfaces and a browser navigation control on web interfaces. Steps that lack valid schema representations or contain unresolvable arguments are discarded with explicitly logged reason codes rather than forced into approximate tool calls.

\begin{figure}[t]
\centering
\includegraphics[width=\linewidth]{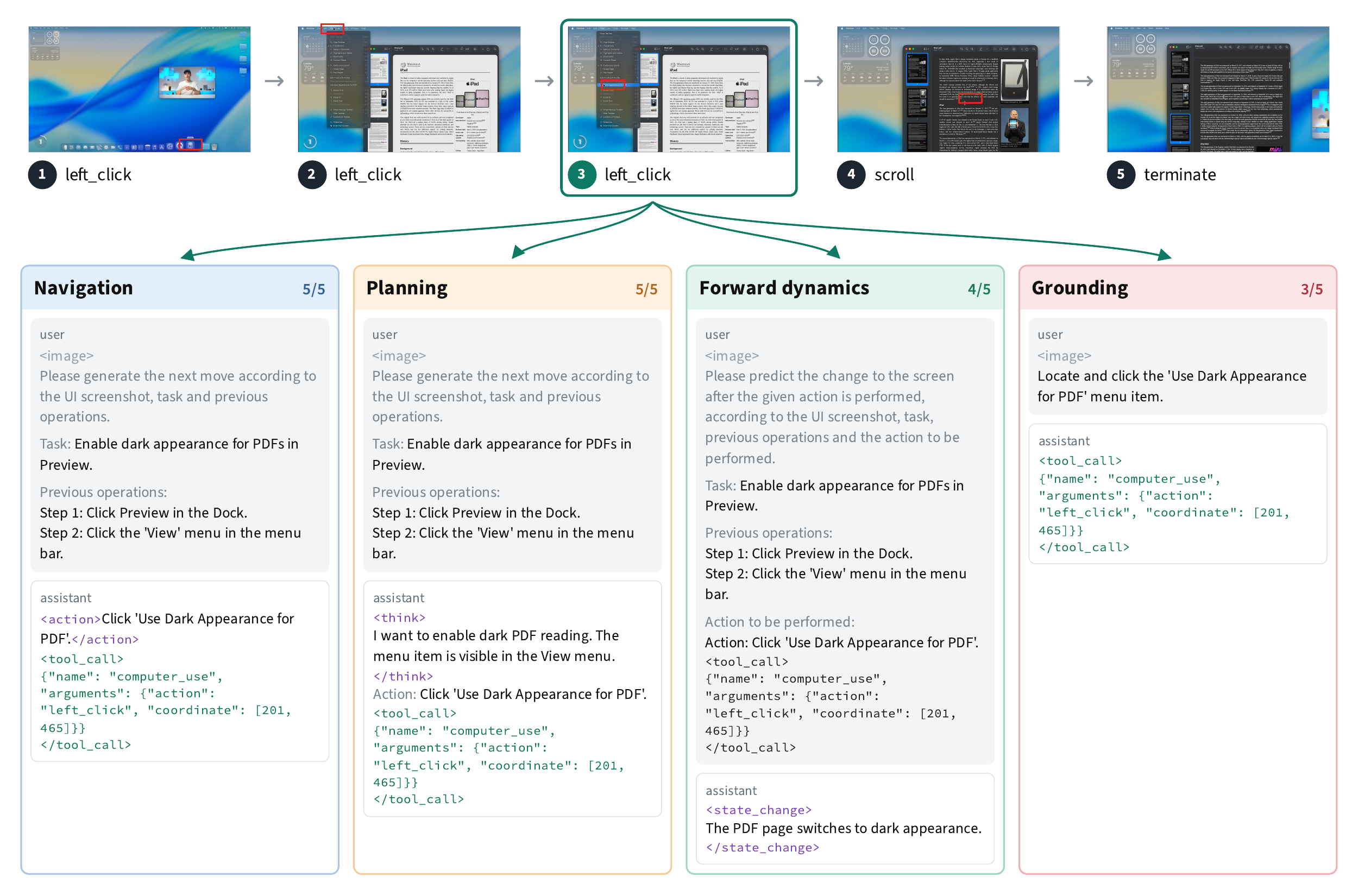}
\caption{\textbf{Multi-task training data conversion.} A single interaction trajectory is transformed into four complementary training views: navigation, planning, forward dynamics, and grounding. The highlighted step illustrates their respective training formats, with gray text indicating shared context, black text indicating step-specific content, and green text indicating the target tool call.}
\label{fig:views}
\end{figure}

\subsubsection{Training-ready Multi-task Conversion}
To construct training-ready datasets, trajectories are decomposed at the step level and formatted into several complementary task views as illustrated in \Cref{fig:views}. Task views that retain sequential context model multi-turn interaction dynamics. Specifically, \textbf{Navigation} tasks predict the next action and tool invocation given the current state and interaction history. \textbf{Planning} tasks extend navigation by generating an explicit reasoning rationale prior to action prediction. \textbf{Forward dynamics} tasks forecast the resulting interface state change produced by a given action. Additionally, navigation includes a \textbf{step-instruction} variant that provides the specific low-level instruction for the current step alongside the task context, shifting the model objective from decision-making to direct instruction execution.

In contrast, single-step \textbf{Grounding} tasks deliberately strip overall task instructions and interaction history. By pairing an isolated screenshot with a textual referring expression, these tasks force the model to resolve visual references purely from image context rather than inferring target positions from prior intent. Furthermore, steps where grounding failed due to target absence are repurposed as negative training examples, explicitly teaching models to decline prediction when targets are missing; these negative examples are retained in the corpus but are not part of the
fine-tuning mixture of \Cref{sec:setup}.

Finally, the conversion pipeline binds system prompts and dialogue templates directly into the output files, producing self-contained and training-ready datasets. In total, the corpus yields 2{,}709{,}896 supervised samples: 787{,}616 navigation, 478{,}688 grounding, 658{,}109 planning and 785{,}483 forward-dynamics samples.
\subsection{Corpus Statistics}
\label{sec:corpus}

\begin{figure*}[!t]
\centering
\includegraphics[width=\textwidth]{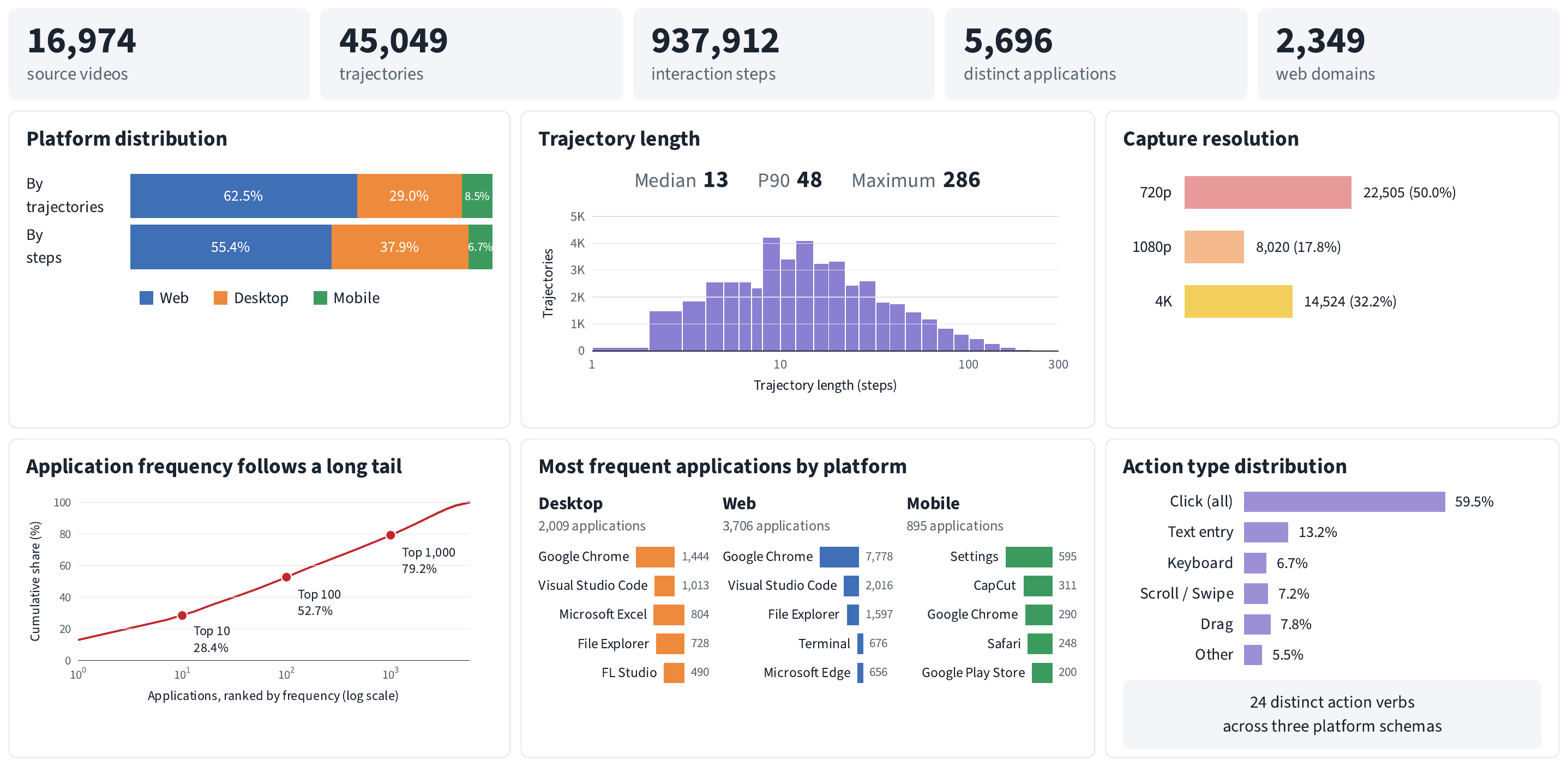}
\caption{\textbf{Overview of the dataset statistics.} The dataset comprises 45,049 trajectories and 937,912 interaction steps from 16,974 source videos, spanning 5,696 applications and 2,349 web domains. The figure summarizes distributions across platforms, trajectory lengths, capture resolutions, applications, and action types.}
\label{fig:corpus_stats}
\end{figure*}

The resulting corpus comprises 45{,}049 trajectories over 16{,}974 source videos, containing 937{,}912 interaction steps; each trajectory additionally carries one synthesized terminal step. The median trajectory is 13 steps long, with a ninetieth-percentile length of 48 and a maximum of 286. The corpus contains both short interactions confined to a single dialog and extended multi-application procedures. \Cref{fig:corpus_stats} summarizes these statistics.

\textbf{Multi-Platform Coverage.} The dataset covers three major computing environments including Web, Desktop, and Mobile, providing multi-platform diversity for GUI agent training. In terms of trajectory proportion, Web constitutes the largest share at 62.5\%, Desktop accounts for 29.0\%, and Mobile represents 8.5\%. Measured by interaction steps, Web comprises 55.4\%, Desktop represents 37.9\%, and Mobile accounts for 6.7\%, reflecting relatively longer interaction sequences within desktop applications. Furthermore, the corpus provides broad multi-resolution coverage, where 720p accounts for 50.0\% at 22,505 trajectories, 4K represents 32.2\% at 14,524 trajectories, and 1080p constitutes the remaining 17.8\% at 8,020 trajectories.

\textbf{Applications and Websites.} Across all computing platforms, the corpus encompasses 5,696 distinct applications and 2,349 web domains. The application distribution varies by platform: Web interactions span 3,706 applications, Desktop encompasses 2,009 applications, and Mobile covers 895 applications (the per-platform counts exceed the total because some applications, such as Google Chrome, occur on more than one platform). On Desktop, the most frequent software includes Google Chrome, Visual Studio Code, Microsoft Excel, File Explorer, and FL Studio. Web workflows center on platforms such as Google Chrome, Visual Studio Code, File Explorer, Terminal, and Microsoft Edge. Mobile workflows focus primarily on utility and consumer applications, led by Settings, CapCut, Google Chrome, Safari, and Google Play Store.

\textbf{Action Space Distribution.} The unified action space defines 24 distinct action verbs managed across three platform-conditioned schemas. The realized interaction distribution is predominantly composed of pointing operations, where clicks of all forms represent 59.5\% of total steps. Text entry accounts for 13.2\%, view manipulations such as scrolling and swiping constitute 7.2\%, and discrete keyboard actions make up 6.7\%. Crucially, drag operations account for 7.8\%, representing a substantial share for an action primitive that is typically scarce in annotation-driven corpora due to its high trajectory labeling complexity. The remaining 5.5\% consists of specialized interaction types.

%% file: sections/05_training.tex
\section{Experimental Setup}
\label{sec:setup}

\subsection{Training}

Supervised fine-tuning is conducted on \textbf{Qwen3-VL-8B-Instruct} \citep{qwen3vl2025} for one epoch using a training mixture of approximately 350,000 samples drawn from the T2T corpus. This mixture is sampled primarily from the 1080p source pool, with mobile tasks supplemented from the 720p and 4K pools to compensate for the relative scarcity of mobile data. The exact mixture composition is detailed in \Cref{tab:recipe} located in \Cref{app:promptcounts}. Across computing platforms, the training mixture allocates approximately 50\% to desktop, 40\% to web, and 10\% to mobile interactions, adjusting the platform distribution relative to the full corpus presented in \Cref{sec:corpus} due to pool-specific sampling. Across task views, the mixture distributes roughly two-thirds of the samples to navigation, and one-third to single-step grounding tasks. The resulting fine-tuned checkpoint, designated as \textbf{T2T-8B}, serves as the primary model evaluated in \Cref{sec:results}.

Training hyperparameters and hardware configurations are summarized in \Cref{tab:hparams}. During fine-tuning, the vision encoder remains frozen while only the multimodal projector and the language model parameters are updated. This design choice relies on two key rationales. First, the pre-trained visual backbone already provides adequate spatial representation for pointing and element localization, allowing the corpus to focus on teaching procedural action execution and tool calling semantics rather than low-level visual recognition. Second, freezing the vision encoder eliminates a primary source of gradient variance caused by optimizing visual features across heterogeneous image resolutions ranging from 720p to 4K.

\begin{table}[t]
\centering\small
\caption{Training configuration.}
\label{tab:hparams}
\begin{tabular}{ll}
\toprule
\textbf{Setting} & \textbf{Value} \\
\midrule
Base model                   & Qwen3-VL-8B-Instruct \\
Objective                    & SFT, loss on assistant turns only \\
Trainable                    & projector $+$ language model; vision encoder frozen \\
Epochs                       & 1 \\
Global batch size            & 48 \\
Learning rate                & $1\times10^{-6}$, cosine schedule, 3\% warmup \\
Gradient clipping            & 5 \\
Precision / memory           & bf16, ZeRO-3, gradient checkpointing \\
Sequence length              & 73{,}728 tokens, with sequence packing \\
Coordinate space             & normalized $1000\times1000$ \\
Image resolution             & native, no downsampling \\
\bottomrule
\end{tabular}
\end{table}

\subsection{Evaluation}

\textbf{Evaluation Dimensions:} To evaluate the capabilities imparted by the T2T corpus, we assess performance across three core dimensions, namely GUI grounding, instruction-guided step execution, and trajectory-level decision-making.

GUI grounding evaluates spatial localization given an instruction and a screenshot. The model predicts normalized coordinates for the target control. Benchmarks include OSWorld-G \citep{xie2025osworldg}, comprising 564 desktop evaluation items alongside its refined-instruction variant, and ScreenSpot-Pro \citep{li2025screenspotpro}, featuring 1,581 items across high-resolution professional software applications. A prediction is scored as correct if the coordinate falls within the ground-truth region according to each benchmark's official evaluator.

Instruction-guided execution measures step-level action generation. Provided with a screenshot, an episode goal, interaction history, and a low-level step instruction, the model generates the complete next tool call. Evaluation is conducted on the official test split of AndroidControl \citep{li2024androidcontrol} in its low-level setting, which isolates step execution from task decomposition; action selection without step instructions is assessed on AgentNetBench below. We report type accuracy, which evaluates whether the predicted action primitive matches, and step success rate, which additionally requires exact agreement across all action parameters. For click actions, predicted points must fall inside bounding boxes extracted from the accessibility tree, falling back to Euclidean distance thresholds only when accessibility boxes are unavailable.

Trajectory-level decision-making assesses multi-step action selection, where models must autonomously determine appropriate action types without explicit step-level guidance. We evaluate on AgentNetBench \citep{xie2025opencua}, an offline desktop benchmark containing 100 trajectories and 1,208 interaction steps. Models receive a single screenshot per step alongside textual action history.

\textbf{Evaluation Protocol:} To eliminate silent evaluation artifacts and pipeline inconsistencies common in benchmark evaluations, we establish strict protocols governing coordinate resolution, prompt fidelity, and inference execution.

Evaluating models under unfamiliar prompt formats often introduces syntax confusion, causing artificial performance drops that reflect formatting errors rather than a lack of visual capability. To prevent this mismatch, we evaluate each model using a prompt structure it was trained for; the prompts are reproduced in \Cref{app:p-eval}. Specifically, on the grounding benchmarks T2T-8B uses the grounding system prompt from its training pipeline, whereas the baseline model uses the computer-use prompt of the official Qwen3-VL cookbook, which requests the target as a \texttt{computer\_use} function call with coordinates in a $1000\times1000$ space. On AndroidControl, both models receive the mobile navigation template from the T2T training pipeline, which presents the \texttt{mobile\_use} tool schema of \Cref{app:actionspace} in the function-calling format of the Qwen-VL agent recipes (tool signatures inside \texttt{<tools>} tags, actions returned inside \texttt{<tool\_call>} tags); under an alternative T2T template that requires a reasoning block before each action, the baseline scored lower (70.32 versus 72.57 step success rate), so the shared template does not disadvantage it.

To ensure valid comparisons, our evaluation pipeline aligns model outputs and benchmark ground truths within a shared coordinate space. While both T2T-8B and Qwen3-VL-8B-Instruct output relative coordinates on a 0 to 1000 scale, individual benchmarks store ground-truth targets in varying formats, such as pixel boxes or normalized ratios. Our pipeline automatically converts model predictions into the format required by each benchmark. For example, to derive pixel positions, relative outputs are scaled against the original screenshot width $W$ and height $H$.

All evaluations are deployed using vLLM with temperature set to zero. Generation token budgets are assigned according to task output requirements. Grounding benchmarks, ScreenSpot-Pro and OSWorld-G, use a limit of 128 tokens for single tool calls. AndroidControl receives a budget of 512 tokens to accommodate action descriptions and arguments. AgentNetBench uses a 1000 token limit, aligning with its official agent configuration.

%% file: sections/07_results.tex
\section{Results}
\label{sec:results}

This section evaluates the effectiveness of the T2T corpus for GUI grounding, instruction-guided execution, and trajectory-level decision-making. \textbf{T2T-8B} denotes Qwen3-VL-8B-Instruct fine-tuned as described in \Cref{sec:setup}, while \textbf{Base} denotes the original checkpoint. As summarized in \Cref{tab:glance}, a single epoch of fine-tuning on approximately 350K tutorial-derived samples improves performance across diverse benchmarks without using in-distribution training data. We further examine performance by target type, software domain, and action primitive, and test whether T2T supervision remains useful alongside in-domain data.

\begin{table}[!htbp]
\centering\small
\caption{Summary of results. All benchmarks are out-of-distribution with
respect to the training data. \emph{Rel.} is the relative improvement over the base model. \emph{Err. red.} is the fraction of the base model's errors removed.}
\label{tab:glance}
\begin{tabular}{llcccc}
\toprule
\textbf{Benchmark} & \textbf{Metric} & \textbf{Base} & \textbf{T2T-8B} & \textbf{Rel.} & \textbf{Err.\ red.} \\
\midrule
OSWorld-G                   & accuracy        & 56.91 & \best{61.52} & \up{8.1\%}  & 10.7\% \\
OSWorld-G\textsubscript{ref}& accuracy        & 65.96 & \best{70.74} & \up{7.2\%}  & 14.0\% \\
ScreenSpot-Pro              & accuracy        & 50.28 & \best{57.87} & \up{15.1\%} & 15.3\% \\
\quad icon targets only     & accuracy        & 20.70 & \best{30.30} & \up{46.4\%} & 12.1\% \\
AndroidControl-Low          & step SR         & 72.57 & \best{80.36} & \up{10.7\%} & 28.4\% \\
AndroidControl-Low          & type accuracy   & 81.60 & \best{91.79} & \up{12.5\%} & 55.4\% \\
\quad \texttt{scroll} steps only & step SR    & 19.08 & \best{64.57} & \up{238\%}  & 56.2\% \\
\bottomrule
\end{tabular}
\end{table}

\subsection{GUI Grounding}

\noindent\textbf{Overall Performance.}
As shown in \Cref{tab:grounding}, T2T-8B improves accuracy across all three grounding evaluations. Accuracy increases from 56.91\% to 61.52\% on OSWorld-G and from 65.96\% to 70.74\% on its refined-instruction variant, OSWorld-G\textsubscript{ref}. The largest absolute gain occurs on ScreenSpot-Pro, where accuracy rises from 50.28\% to 57.87\% (+7.59 percentage points). These results support the effectiveness of tutorial-derived trajectories for GUI grounding, particularly on visually demanding interfaces. A limitation remains on target-absent cases: T2T-8B correctly rejects none of the 54 such items in OSWorld-G, compared with three for Base. This is consistent with the exclusion of negative grounding examples from the fine-tuning mixture (\Cref{sec:actionspace}). On the remaining 510 items, accuracy improves from 62.35\% to 68.04\%.

\begin{table}[!htbp]
\centering\small
\caption{GUI grounding accuracy (\%). OSWorld-G\textsubscript{ref} is the
refined-instruction variant: the same 564 items and identical target regions,
with 507 of the instructions rewritten from functional intent into explicit
appearance-and-position descriptions.}
\label{tab:grounding}
\begin{tabular}{lccc}
\toprule
\textbf{Model} & \textbf{OSWorld-G} & \textbf{OSWorld-G\textsubscript{ref}} & \textbf{ScreenSpot-Pro} \\
& \small(564) & \small(564) & \small(1{,}581) \\
\midrule
Base     & 56.91 & 65.96 & 50.28 \\
T2T-8B   & \best{61.52} & \best{70.74} & \best{57.87} \\
\midrule
$\Delta$ & \up{4.61} & \up{4.78} & \up{7.59} \\
\bottomrule
\end{tabular}
\end{table}

\noindent\textbf{Performance by Target Type.}
\Cref{tab:sspro} breaks down ScreenSpot-Pro performance by target type and software domain. T2T-8B improves all twelve domain-by-type subsets. Accuracy on text targets increases from 68.58\% to 74.92\% (+6.34 points), while icon-target accuracy rises from 20.70\% to 30.30\% (+9.60 points, or 46.4\% relative). The larger gain on icons, which offer fewer explicit textual cues, suggests that T2T supervision is particularly useful for visually grounded element localization.

\noindent\textbf{Performance by Software Domain.}
Gains extend across all six domains, with the largest absolute improvements in CAD (+11.88 points) and OS interfaces (+10.20 points). In OS interfaces, icon accuracy increases from 19.10\% to 34.83\%. In CAD, both text targets (48.22\% to 61.42\%) and icon targets (15.62\% to 23.44\%) improve substantially. Together, these results highlight the value of tutorial-derived supervision for precise grounding in complex interfaces with dense controls and limited textual cues.

\begin{table}[!htbp]
\centering\small
\caption{ScreenSpot-Pro accuracy (\%) by domain and target type. \emph{Every}
one of the twelve cells improves, as do all six domain totals and both target
types.}
\label{tab:sspro}
\begin{tabular}{lrcc@{\hskip 1.2em}cc@{\hskip 1.2em}cc}
\toprule
& & \multicolumn{2}{c}{\textbf{Text}} & \multicolumn{2}{c}{\textbf{Icon}} & \multicolumn{2}{c}{\textbf{All}} \\
\cmidrule(lr){3-4}\cmidrule(lr){5-6}\cmidrule(lr){7-8}
\textbf{Domain} & \textbf{$n$} & Base & T2T & Base & T2T & Base & T2T \\
\midrule
Office     & 230 & 77.40 & \best{80.23} & 32.08 & \best{43.40} & 66.96 & \best{71.74} \\
Scientific & 254 & 78.47 & \best{82.64} & 26.36 & \best{37.27} & 55.91 & \best{62.99} \\
OS         & 196 & 71.03 & \best{76.64} & 19.10 & \best{34.83} & 47.45 & \best{57.65} \\
Dev        & 299 & 77.27 & \best{80.52} & 20.00 & \best{29.66} & 49.50 & \best{55.85} \\
CAD        & 261 & 48.22 & \best{61.42} & 15.62 & \best{23.44} & 40.23 & \best{52.11} \\
Creative   & 341 & 65.66 & \best{72.73} & 16.08 & \best{20.98} & 44.87 & \best{51.03} \\
\midrule
\textbf{All} & \textbf{1{,}581} & 68.58 & \best{74.92} & 20.70 & \best{30.30} & 50.28 & \best{57.87} \\
\midrule
$\Delta$ & & \up{6.34} & & \up{9.60} & & \up{7.59} & \\
\bottomrule
\end{tabular}
\end{table}

\subsection{Instruction-guided Execution}

\noindent\textbf{Overall Performance.}
We next evaluate step-level execution on the official AndroidControl low-level test split, where the model receives a screenshot, task goal, action history, and step instruction and must generate the complete next tool call. This is an out-of-domain evaluation: no AndroidControl training examples are used for T2T-8B, and mobile data accounts for only approximately 10\% of its fine-tuning mixture. Across 7,846 test steps (\Cref{tab:aclow}), step success rate increases from 72.57\% to 80.36\% (+7.79 points), while action-type accuracy rises from 81.60\% to 91.79\% (+10.19 points). These improvements remove 28.4\% of baseline step errors and 55.4\% of action-type errors. Gains also hold on unseen applications (+6.73 points), tasks (+6.06 points), and categories (+5.94 points), indicating transfer beyond the in-distribution split.

\begin{table}[!htbp]
\centering\small
\caption{AndroidControl, low-level setting, official test split (7{,}846
steps). Both models are evaluated under the same prompt template and the same
element-box scoring criterion. IDD = in-distribution split.}
\label{tab:aclow}
\begin{tabular}{lcc@{\hskip 2em}cccc}
\toprule
& & & \multicolumn{4}{c}{\textbf{Step SR by split}} \\
\cmidrule(lr){4-7}
\textbf{Model} & \textbf{Type Acc} & \textbf{Step SR} & IDD & App\tsub{unseen} & Task\tsub{unseen} & Cat.\tsub{unseen} \\
\midrule
Base   & 81.60 & 72.57 & 73.33 & 71.52 & 71.86 & 70.99 \\
T2T-8B & \best{91.79} & \best{80.36} & \best{83.00} & \best{78.25} & \best{77.92} & \best{76.93} \\
\midrule
$\Delta$ & \up{10.19} & \up{7.79} & \up{9.67} & \up{6.73} & \up{6.06} & \up{5.94} \\
\bottomrule
\end{tabular}
\end{table}

\noindent\textbf{Performance by Action Type.}
As shown in \Cref{tab:acaction}, the largest gain occurs on \texttt{scroll}, whose step success rate increases from 19.08\% to 64.57\% (+45.49 points). Scroll and swipe operations are well represented in tutorial videos (40{,}255 native \texttt{scroll} and 26{,}675 \texttt{swipe} steps; \Cref{tab:nativefull}), providing supervision for an action that is difficult to annotate manually. Gains are also observed for \texttt{open\_app} (+11.67 points) and \texttt{input\_text} (+6.81 points). The \texttt{long\_press} gain (+22.22 points) is promising but is measured on only nine examples. Click performance changes modestly (89.52\% to 88.58\%), whereas \texttt{wait} declines from 23.63\% to 20.46\%. The latter remains challenging because a single screenshot provides limited temporal evidence. Overall, the results point to better coverage of non-click action primitives without a large loss on the dominant click action.

\begin{table}[!htbp]
\centering\small
\caption{AndroidControl-Low step success rate (\%) by ground-truth action type.}
\label{tab:acaction}
\begin{tabular}{lrccr}
\toprule
\textbf{Action} & \textbf{$n$} & \textbf{Base} & \textbf{T2T-8B} & \textbf{$\Delta$} \\
\midrule
\texttt{scroll}        & 1{,}211 & 19.08 & \best{64.57} & \up{45.49} \\
\texttt{long\_press}   & \phantom{1{,}21}9 & 66.67 & \best{88.89} & \up{22.22} \\
\texttt{open\_app}     & \phantom{1{,}}608 & 73.36 & \best{85.03} & \up{11.67} \\
\texttt{input\_text}   & \phantom{1{,}}632 & 88.13 & \best{94.94} & \up{6.81} \\
\texttt{navigate\_back}& \phantom{1{,}}343 & 91.25 & \best{92.42} & \up{1.17} \\
\texttt{click}         & 4{,}476 & \best{89.52} & 88.58 & $-0.94$ \\
\texttt{wait}          & \phantom{1{,}}567 & \best{23.63} & 20.46 & $-3.17$ \\
\bottomrule
\end{tabular}
\end{table}

\subsection{Complementarity with In-domain Data}
\label{sec:indomain}

\noindent\textbf{Experimental Setup.}
To test whether tutorial-derived supervision remains useful when target-domain data is available, we construct an approximately 6K-sample set from the AndroidControl training split using the conversion pipeline in \Cref{sec:pipeline}. We then perform one additional SFT epoch starting from either Base or T2T-8B. Because downward scrolls dominate the original data (78.5\%), scroll examples are resampled to a more balanced direction distribution (45/25/18/12\% for down/up/right/left; \Cref{app:scrolldir}). This setup helps distinguish improvements in choosing to scroll from improvements in predicting its direction.

\noindent\textbf{Complementary Gains.}
\Cref{tab:indomain} shows that T2T-8B, without AndroidControl training data, reaches 80.36\% step success rate, compared with 72.57\% for Base and 84.08\% for Base fine-tuned on the 6K in-domain samples. Combining the two sources performs best: T2T-8B followed by in-domain fine-tuning reaches 86.01\%, outperforming Base with the same in-domain fine-tuning by 1.93 points. Thus, tutorial-derived data contributes beyond what this modest amount of in-domain supervision provides.

\begin{table}[!htbp]
\centering\small
\caption{Complementarity with in-domain data, AndroidControl low-level setting.
Same split and criterion as \Cref{tab:aclow}; the last four columns are step
success rate by action type.}
\label{tab:indomain}
\begin{tabular}{lcc@{\hskip 1.5em}cccc}
\toprule
\textbf{Model} & \textbf{Type Acc} & \textbf{Step SR} & \texttt{scroll} & \texttt{open\_app} & \texttt{input\_text} & \texttt{click} \\
\midrule
Base                     & 81.60 & 72.57 & 19.08 & 73.36 & 88.13 & \best{89.52} \\
Base $+$ 6K in-domain    & 94.47 & 84.08 & 84.56 & 88.49 & 93.99 & 83.71 \\
T2T-8B                   & 91.79 & 80.36 & 64.57 & 85.03 & \best{94.94} & 88.58 \\
T2T-8B $+$ 6K in-domain  & \best{95.32} & \best{86.01} & \best{86.87} & \best{88.98} & 94.62 & 86.15 \\
\bottomrule
\end{tabular}
\end{table}

\noindent\textbf{Scroll Direction Analysis.}
The additional in-domain epoch raises T2T-8B's overall step success rate by 5.65 points, with notable gains on \texttt{scroll} and \texttt{open\_app}. For scroll steps whose action type is already predicted correctly, direction accuracy rises from 67.9\% to 87.8\%. This indicates that the additional supervision improves the direction argument, rather than only the decision to scroll. The test split contains 78.2\% downward scrolls, but T2T-8B initially predicts \texttt{down} on only 65.8\% of test scroll steps. After the in-domain epoch, this prediction rate becomes 79.2\%, close to the test distribution despite the more balanced training directions. This pattern is consistent with improved use of screen content to predict direction, although matching aggregate direction frequencies alone does not establish the underlying mechanism. Further distributions are provided in \Cref{app:scrolldir}.

\subsection{Trajectory-level Evaluation}
\label{sec:agentnet-heldout}

\noindent\textbf{Evaluation Protocol.}
Finally, we evaluate action selection without step-level instructions on AgentNetBench, an offline desktop benchmark containing 100 trajectories and 1{,}208 steps. Both models receive one screenshot per step and the preceding action history as text. T2T-8B uses the desktop navigation template from its training pipeline, while Base uses the benchmark's Qwen-VL agent prompt with its coordinate convention adapted to the same normalized $0$--$1000$ space. The prompts are provided in \Cref{app:p-eval}.

\begin{table}[!htbp]
\centering\small
\caption{AgentNetBench overall score (\%).}
\label{tab:anb-overall}
\begin{tabular}{lcc}
\toprule
\textbf{Model} & \textbf{Score} \\
\midrule
Base     & 63.56 \\
T2T-8B   & \best{65.23} \\
\bottomrule
\end{tabular}
\end{table}

\noindent\textbf{Overall and Per-action Results.}
T2T-8B improves the overall score from 63.56\% to 65.23\% (\Cref{tab:anb-overall}). The per-action breakdown in \Cref{tab:anb-actions} shows higher scores on seven of ten reported action types. Particularly large differences appear for \texttt{doubleClick} (11.76\% to 70.59\%), \texttt{terminate} (0.00\% to 21.65\%), and \texttt{moveTo} (13.16\% to 31.21\%). These gains suggest broader support for GUI-specific operations, although some categories contain relatively few evaluation examples. Performance declines slightly on \texttt{click} and \texttt{write}, and more noticeably on \texttt{hotkey}. The modest aggregate improvement therefore indicates progress in action selection while leaving substantial room for stronger trajectory-level decision-making.

\begin{table}[!htbp]
\centering\small
\caption{AgentNetBench per-action score (\%) under the single-image protocol,
ordered by margin. $n$ is the number of ground-truth actions of each type; a step may contain more than one action (e.g., \texttt{moveTo} followed by \texttt{dragTo}), so the counts sum to 1{,}269 over the 1{,}208 steps.}
\label{tab:anb-actions}
\begin{tabular}{lrccr}
\toprule
\textbf{Action} & \textbf{$n$} & \textbf{Base} & \textbf{T2T-8B} & \textbf{$\Delta$} \\
\midrule
\texttt{doubleClick} & 20  & 11.76 & \best{70.59} & \up{58.8} \\
\texttt{terminate}   & 97  & \phantom{0}0.00 & \best{21.65} & \up{21.7} \\
\texttt{moveTo}      & 45  & 13.16 & \best{31.21} & \up{18.1} \\
\texttt{rightClick}  & 17  & 14.29 & \best{21.43} & \up{7.1} \\
\texttt{scroll}      & 18  & \phantom{0}4.62 & \best{11.33} & \up{6.7} \\
\texttt{dragTo}      & 27  & \phantom{0}4.35 & \best{\phantom{0}7.70} & \up{3.4} \\
\texttt{press}       & 28  & 50.00 & \best{52.63} & \up{2.6} \\
\midrule
\texttt{click}       & 850 & \best{74.99} & 73.44 & $-1.6$ \\
\texttt{write}       & 137 & \best{74.24} & 72.51 & $-1.7$ \\
\texttt{hotkey}      & 30  & \best{26.67} & 16.67 & $-10.0$ \\
\bottomrule
\end{tabular}
\end{table}

%% file: sections/09_conclusion.tex
\section{Conclusion}

We present T2T, a fully automated pipeline for synthesizing GUI interaction trajectories from unannotated video tutorials across desktop, web, and mobile platforms. Through video filtering, action extraction, presence-aware spatial grounding, and action normalization, T2T produces over one million training samples without human annotation. Fine-tuning Qwen3-VL-8B on a subset of these samples yields consistent improvements across diverse GUI benchmarks without using in-distribution training data, demonstrating the potential of tutorial-derived trajectories as a scalable supervision source for generalist GUI agents.

\textbf{Limitations.}
T2T dataset quality is assessed primarily through automated verification rather than systematic human annotation audits. Moreover, negative grounding examples are excluded from the current training mixture, limiting the model's ability to abstain when targets are absent. Current evaluation focuses on step-level and offline trajectory-level benchmarks, leaving online task completion for future investigation. Mobile tutorial videos introduce visual challenges, including background clutter, occlusion, and perspective distortion in externally captured recordings.

%% file: sections/99_appendix.tex
\FloatBarrier
\section{Video Filtering Parameters}
\label{app:filterparams}

\Cref{tab:filterparams} summarizes the thresholds used for video filtering and preprocessing (\Cref{sec:filtering}). All values were selected on a development sample and held fixed across source pools.

\begin{table}[ht]
\centering\small
\caption{Video preprocessing and filtering parameters.}
\label{tab:filterparams}
\begin{tabularx}{\textwidth}{lX}
\toprule
\textbf{Filter} & \textbf{Setting} \\
\midrule
Cursor detector & single-class YOLO on cursor glyphs; 1280\,px input; confidence threshold 0.3 \\
Probe stride & every 30th frame (one probe per second at 30\,fps); every 10th frame for the 720p pool \\
Static-frame skip & detector skipped when mean absolute intensity difference to the last cursor-bearing frame $<10$ \\
Segment gap tolerance & segment closes after 2\,s without a detection \\
Minimum segment length & 6\,s \\
Desktop/web retention & union of active segments $\geq 30\%$ of video duration \\
Mobile visual filter & CLIP zero-shot on 8 uniformly sampled frames; $\geq 4$ classified as mobile UI \\
Mobile linguistic filter & $\geq 2$ imperative instruction phrases in the transcript, plus a normalized instruction-density threshold \\
Annotated footage & 16 sampled frames; reject if colored box outline in $\geq 25\%$ of frames, or mean annotation-color coverage $> 15\%$ of frame area \\
\bottomrule
\end{tabularx}
\end{table}

\FloatBarrier
\section{Action Space and Vocabulary}
\label{app:actionspace}

As described in \Cref{sec:actionspace}, action normalization preserves the extractor's native outputs before mapping them to platform-specific tool calls. This appendix specifies the native vocabulary, the three tool schemas, and the observed mappings.

\subsection{Stage 1: The Native Extractor Vocabulary}
\label{app:fulltables}

\Cref{tab:nativefull} reports the native extractor vocabulary, including synonym groups merged during normalization. The original action labels remain available in the trajectory logs.

\begin{table}[htbp]
\centering\small
\caption{Native extractor action vocabulary over all candidate steps. Synonym groups that are merged during normalization are marked.}
\label{tab:nativefull}
\begin{tabular}{lrrl}
\toprule
\textbf{Action} & \textbf{Steps} & \textbf{Share} & \textbf{Note} \\
\midrule
\texttt{left\_click}      & 384{,}160 & 40.96\% & \multirow{2}{*}{$\Big\}$ merged $\rightarrow$ click family} \\
\texttt{click}            & 152{,}424 & 16.25\% & \\
\texttt{type}             & 117{,}332 & 12.51\% & \multirow{2}{*}{$\Big\}$ merged $\rightarrow$ \texttt{type}} \\
\texttt{write}            & \phantom{00}6{,}281 & \phantom{0}0.67\% & \\
\texttt{left\_click\_drag}& \phantom{0}73{,}587 & \phantom{0}7.85\% & \\
\texttt{key}              & \phantom{0}45{,}540 & \phantom{0}4.86\% & \multirow{2}{*}{$\Big\}$ merged $\rightarrow$ \texttt{key}} \\
\texttt{hotkey}           & \phantom{0}17{,}657 & \phantom{0}1.88\% & \\
\texttt{scroll}           & \phantom{0}40{,}255 & \phantom{0}4.29\% & \\
\texttt{wait}             & \phantom{0}34{,}785 & \phantom{0}3.71\% & \\
\texttt{swipe}            & \phantom{0}26{,}675 & \phantom{0}2.84\% & intent-resolved \\
\texttt{right\_click}     & \phantom{0}14{,}033 & \phantom{0}1.50\% & \\
\texttt{double\_click}    & \phantom{00}6{,}735 & \phantom{0}0.72\% & \\
\texttt{select\_option}   & \phantom{00}5{,}240 & \phantom{0}0.56\% & \\
\texttt{mouse\_move}      & \phantom{00}2{,}845 & \phantom{0}0.30\% & \\
\texttt{select}           & \phantom{00}1{,}936 & \phantom{0}0.21\% & \\
\texttt{open\_url}        & \phantom{00}1{,}852 & \phantom{0}0.20\% & \\
\texttt{open\_app}        & \phantom{00}1{,}804 & \phantom{0}0.19\% & \\
\texttt{navigate\_back}   & \phantom{00}1{,}000 & \phantom{0}0.11\% & \multirow{4}{*}{$\Bigg\}$ merged $\rightarrow$ back / system button} \\
\texttt{navigate\_home}   & \phantom{000}\phantom{0}843 & \phantom{0}0.09\% & \\
\texttt{go\_backward}     & \phantom{000}\phantom{0}506 & \phantom{0}0.05\% & \\
\texttt{go\_back}         & \phantom{000}\phantom{0}442 & \phantom{0}0.05\% & \\
\texttt{long\_press}      & \phantom{000}\phantom{0}774 & \phantom{0}0.08\% & \\
\texttt{hscroll}          & \phantom{000}\phantom{0}508 & \phantom{0}0.05\% & \\
\texttt{system\_button}   & \phantom{000}\phantom{0}450 & \phantom{0}0.05\% & \\
\texttt{open}             & \phantom{000}\phantom{0}200 & \phantom{0}0.02\% & \\
\texttt{go\_forward}      & \phantom{0000}\phantom{00}28 & $<$0.01\% & \\
\texttt{triple\_click}    & \phantom{0000}\phantom{00}12 & $<$0.01\% & \\
\texttt{middle\_click}    & \phantom{0000}\phantom{000}8 & $<$0.01\% & \\
\bottomrule
\end{tabular}
\end{table}

\subsection{Stage 2: The Platform-conditioned Tool-call Schemas}

The schemas below define valid actions and arguments for each platform. ``Grd'' marks actions whose coordinates identify a spatial target rather than a scroll region or direction.

\begin{table}[htbp]
\centering\small
\caption{\texttt{mobile\_use} (12 actions).}
\label{tab:mobileuse}
\begin{tabular}{llc}
\toprule
\textbf{Action} & \textbf{Parameters} & \textbf{Grd} \\
\midrule
\texttt{click}         & \texttt{coordinate} & \checkmark \\
\texttt{long\_press}   & \texttt{coordinate}, \texttt{time} & \checkmark \\
\texttt{scroll}        & \texttt{coordinate}, \texttt{direction}, \texttt{amount} & \\
\texttt{swipe}         & \texttt{coordinate}, \texttt{coordinate2} & \\
\texttt{type}          & \texttt{text} & \\
\texttt{system\_button}& \texttt{button} $\in$ \{Back, Home, Menu, Enter\} & \\
\texttt{open}          & \texttt{text} & \\
\texttt{key}           & \texttt{text} (adb keyevent syntax) & \\
\texttt{wait}          & \texttt{time} & \\
\texttt{answer}        & \texttt{text} & \\
\texttt{interact}      & \texttt{text} & \\
\texttt{terminate}     & \texttt{status} $\in$ \{success, failure\} & \\
\bottomrule
\end{tabular}
\end{table}

\begin{table}[htbp]
\centering\small
\caption{\texttt{computer\_use} (desktop, 15 actions) and \texttt{browser\_use}
(web, 19 actions). The web schema is the desktop schema plus the four actions
marked \emph{web}.}
\label{tab:computeruse and browseruse}
\begin{tabular}{llcl}
\toprule
\textbf{Action} & \textbf{Parameters} & \textbf{Grd} & \\
\midrule
\texttt{left\_click}       & \texttt{coordinate} & \checkmark & \\
\texttt{right\_click}      & \texttt{coordinate} & \checkmark & \\
\texttt{middle\_click}     & \texttt{coordinate} & \checkmark & \\
\texttt{double\_click}     & \texttt{coordinate} & \checkmark & \\
\texttt{triple\_click}     & \texttt{coordinate} & \checkmark & \\
\texttt{mouse\_move}       & \texttt{coordinate} & \checkmark & \\
\texttt{left\_click\_drag} & \texttt{coordinate}, \texttt{coordinate2} & \checkmark & \\
\texttt{scroll}            & \texttt{coordinate}, \texttt{clicks} & & \\
\texttt{hscroll}           & \texttt{coordinate}, \texttt{clicks} & & \\
\texttt{type}              & \texttt{text} & & \\
\texttt{key}               & \texttt{keys} & & \\
\texttt{wait}              & \texttt{time} & & \\
\texttt{terminate}         & \texttt{status} & & \\
\texttt{answer}            & \texttt{text} & & \\
\texttt{interact}          & \texttt{text} & & \\
\midrule
\texttt{select}            & \texttt{coordinate}, \texttt{text} & \checkmark & \emph{web} \\
\texttt{open\_url}         & \texttt{text} & & \emph{web} \\
\texttt{go\_back}          & --- & & \emph{web} \\
\texttt{go\_forward}       & --- & & \emph{web} \\
\bottomrule
\end{tabular}
\end{table}

\paragraph{Direction convention.}
In \texttt{mobile\_use.scroll}, \texttt{direction} denotes the direction of movement through content, not finger motion; \texttt{down} reveals content below the visible region. Desktop and web scrolling follow the same convention, with positive \texttt{clicks} moving toward the top. This convention is shared by the corpus, training prompts, and evaluation.

\subsection{Native-to-tool-call Mapping}

\Cref{tab:mappingfull} summarizes native-to-schema conversion, including actions that could not be mapped to a valid tool call.

\begin{table}[htbp]
\centering\small
\caption{Native action $\rightarrow$ tool-call mapping, measured over the
720p and 4K source pools; native verbs that do not occur in these pools are omitted. ``Unconverted'' counts steps for which no defensible tool call could be
produced; overall, $97.0\%$ of steps map successfully.}
\label{tab:mappingfull}
\begin{tabular}{llr}
\toprule
\textbf{Native action} & \textbf{Mapped to (counts)} & \textbf{Unconv.} \\
\midrule
\texttt{left\_click}      & \texttt{left\_click} 251{,}099; \texttt{click} 2{,}143 & 1{,}746 \\
\texttt{click}            & \texttt{left\_click} 107{,}985; \texttt{click} 36{,}479 & 1{,}038 \\
\texttt{type}             & \texttt{type} 81{,}991 & \phantom{0,}142 \\
\texttt{left\_click\_drag}& \texttt{left\_click\_drag} 49{,}312; \texttt{swipe} 420 & 7{,}074 \\
\texttt{wait}             & \texttt{wait} 23{,}974 & \phantom{0,0}53 \\
\texttt{key}              & \texttt{key} 22{,}426 & \phantom{0,0}28 \\
\texttt{scroll}           & \texttt{scroll} 20{,}120 & \phantom{0,}279 \\
\texttt{swipe}            & \texttt{scroll} 13{,}360; \texttt{swipe} 6{,}277; \texttt{left\_click\_drag} 331 & 5{,}079 \\
\texttt{hotkey}           & \texttt{key} 17{,}609; \texttt{system\_button} 3 & \phantom{0,0}45 \\
\texttt{right\_click}     & \texttt{right\_click} 9{,}794 & \phantom{0,0}61 \\
\texttt{write}            & \texttt{type} 6{,}258 & \phantom{0,0}23 \\
\texttt{select\_option}   & \texttt{select} 3{,}518; \texttt{left\_click} 1{,}683; \texttt{click} 17 & \phantom{0,0}22 \\
\texttt{double\_click}    & \texttt{double\_click} 4{,}361 & \phantom{0,0}40 \\
\texttt{open\_app}        & \texttt{open} 1{,}177 & \phantom{0,}627 \\
\texttt{open\_url}        & \texttt{open\_url} 936 & \phantom{0,}321 \\
\texttt{navigate\_home}   & \texttt{system\_button} 827 & \phantom{0,0}16 \\
\texttt{navigate\_back}   & \texttt{system\_button} 752; \texttt{go\_back} 106 & \phantom{0,}142 \\
\texttt{long\_press}      & \texttt{long\_press} 556 & \phantom{0,0}91 \\
\texttt{go\_backward}     & \texttt{go\_back} 457; \texttt{system\_button} 8 & \phantom{0,0}41 \\
\texttt{triple\_click}    & \texttt{triple\_click} 11 & \phantom{0,00}0 \\
\texttt{go\_forward}      & \texttt{go\_forward} 10 & \phantom{0,00}0 \\
\midrule
\multicolumn{2}{l}{\textbf{Mapping rate}} & \textbf{97.0\%} \\
\bottomrule
\end{tabular}
\end{table}

\FloatBarrier
\section{Per-Step Provenance Schema}
\label{app:provenance}

Each reconstructed step records extraction, frame selection, grounding, and action-mapping decisions, as detailed in the table below. These provenance fields support filtering and auditing without rerunning the pipeline.

\begin{table}[htbp]
\centering\small
\caption{Recorded fields per step. \emph{Ext} = written by the extraction model,
\emph{Sel} = observation-frame selector, \emph{Grd} = grounder, \emph{Map} =
action-space mapper.}
\begin{tabularx}{\textwidth}{llX}
\toprule
\textbf{Field} & \textbf{By} & \textbf{Meaning} \\
\midrule
\pth{action\_type}                    & Ext & Native verb (\Cref{tab:nativefull}). \\
\pth{global\_start/end\_time}         & Ext & Action window over the source video. \\
\pth{timestamp\_anchored}             & Ext & Whether the window was pinned to a detected event. \\
\pth{keyframes.start/end\_frame}      & Ext & Pre-action and post-action frames. \\
\pth{transcripts.before/during/after} & Ext & Narration windows around the action. \\
\pth{target}                          & Ext & Natural-language name of the acted-on element. \\
\pth{drag\_end\_target}               & Ext & Destination name, drags only. \\
\pth{inner\_monologue.thought}        & Ext & First-person rationale, contemporaneous. \\
\pth{inner\_monologue.action\_description} & Ext & One-sentence imperative; becomes the \pth{<action>} turn. \\
\pth{state\_change}                   & Ext & Expected post-condition. \\
\pth{parsed\_actions}                 & Ext & Raw parameters before normalization. \\
\pth{resized\_width/height}           & Ext & Frame dimensions the coordinates are relative to. \\
\midrule
\pth{observation\_selection.method}   & Sel & Pre-action frame, an earlier frame, or a cursor-guarded frame. \\
\pth{observation\_selection.reason}   & Sel & Why that frame was chosen. \\
\pth{observation\_selection.t3\_target\_xy} & Sel & Cursor position used by the guard. \\
\midrule
\pth{grounding.source}, \pth{.tool}   & Grd & Grounder identity. \\
\pth{grounding.mode}                  & Grd & \pth{point} or \pth{bbox}. \\
\pth{grounding.query}                 & Grd & The phrase submitted to the grounder; equal to \pth{target} unless rewritten. \\
\pth{grounding.qc.present}            & Grd & False marks an \emph{absent} verdict. \\
\pth{grounding.qc.confidence}         & Grd & Self-reported confidence of the grounder. \\
\pth{grounding.qc.frames\_tried}      & Grd & Number of frames attempted. \\
\pth{grounding.qc.escalated}          & Grd & Whether a second (verification-path) attempt was made. \\
\pth{grounding.qc.verified}           & Grd & Result of the verification pass, where run. \\
\pth{grounding.point\_norm/\_px}      & Grd & Coordinate, normalized and in pixels. \\
\midrule
\pth{tool\_call.action}, \pth{.arguments} & Map & Emitted tool call. \\
\pth{tool\_call.ok}                     & Map & Whether a defensible call could be produced. \\
\pth{tool\_call.backfilled}             & Map & Parameters recovered from a legacy field. \\
\pth{uncertainty\_codes}              & Map & Uncertainty codes raised during mapping. \\
\bottomrule
\end{tabularx}
\end{table}

A compact per-step index retains the fields required for corpus statistics and can be joined to trajectories by video, task, and timestamp.

\FloatBarrier
\section{Corpus Samples}
\label{app:sample}

The following examples reproduce the stored training format. The first is a mobile navigation sample:

\begin{lstlisting}
{
  "image":  "-3msSQJ4_v8/task_0/step_01_start.jpg",
  "width":  3840,
  "height": 2160,
  "conversations": [
    {"from": "human", "value":
      "<image>\nPlease generate the next move according to the UI screenshot,
       task and previous operations.\n\n
       Task: Update or open TikTok, browse sticker comments, and create a new
       live sticker from a video.\n\n
       Previous operations:\nStep 1: Scroll down on the comments list."},
    {"from": "gpt", "value":
      "<action>Scroll down on the comments list again.</action>\n
       <tool_call>\n
       {\"name\": \"mobile_use\", \"arguments\": {\"action\": \"scroll\",
        \"coordinate\": [498, 536], \"direction\": \"down\"}}\n
       </tool_call>"}
  ]
}
\end{lstlisting}

The second is a desktop grounding sample, without task-level context or action history:

\begin{lstlisting}
{
  "image":  "--LtGxJ8U4E/step_00_start.jpg",
  "width":  1920,
  "height": 1080,
  "conversations": [
    {"from": "human", "value":
      "<image>\nClick the '+ Create project' button in the large cyan banner."},
    {"from": "gpt", "value":
      "<tool_call>\n
       {\"name\": \"computer_use\", \"arguments\": {\"action\": \"left_click\",
        \"coordinate\": [578, 123]}}\n
       </tool_call>"}
  ]
}
\end{lstlisting}

\FloatBarrier
\section{Training Mixture}
\label{app:promptcounts}
\Cref{tab:recipe} details the training mixture used for supervised fine-tuning (\Cref{sec:setup}).

\begin{table}[htbp]
\centering\small
\caption{Fine-tuning mixture. \emph{Nav} = navigation, \emph{Grd} = grounding,
\emph{SI} = step-instruction navigation. \emph{Source} gives the capture
resolution of the pool each subset is drawn from; the two 4K pools are reported
together.}
\label{tab:recipe}
\begin{tabular}{llllr}
\toprule
\textbf{Subset} & \textbf{View} & \textbf{Source} & \textbf{Prompt template} & \textbf{Samples} \\
\midrule
Desktop navigation      & Nav & 1080p & Desktop navigation & 108{,}406 \\
Web navigation          & Nav & 1080p & Web navigation     & \phantom{0}89{,}145 \\
Desktop grounding       & Grd & 1080p & Desktop grounding  & \phantom{0}59{,}938 \\
Web grounding           & Grd & 1080p & Web grounding      & \phantom{0}50{,}535 \\
Mobile navigation, SI   & SI  & 720p  & Mobile navigation  & \phantom{0}13{,}279 \\
Mobile navigation       & Nav & 1080p & Mobile navigation  & \phantom{00}6{,}894 \\
Mobile navigation, SI   & SI  & 1080p & Mobile navigation  & \phantom{00}6{,}827 \\
Mobile navigation, SI   & SI  & 4K    & Mobile navigation  & \phantom{00}5{,}102 \\
Mobile grounding        & Grd & 1080p & Mobile grounding   & \phantom{00}4{,}881 \\
Desktop drag grounding  & Grd & 1080p & Desktop grounding  & \phantom{000,}401 \\
Web drag grounding      & Grd & 1080p & Web grounding      & \phantom{000,}177 \\
\midrule
\multicolumn{4}{l}{\textbf{Total}} & \textbf{345{,}585} \\
\bottomrule
\end{tabular}
\end{table}

\FloatBarrier
\section{Prompts}
\label{app:prompts}

This appendix reproduces the pipeline prompts and their roles (\Cref{tab:promptmap}). Training-time system prompts are generated from the same action-schema registry used for normalization.

\begin{table}[htbp]
\centering\small
\caption{Prompt roles across pipeline stages.}
\label{tab:promptmap}
\begin{tabular}{llll}
\toprule
\textbf{Stage} & \textbf{Model} & \textbf{Purpose} & \textbf{Prompt} \\
\midrule
Filtering   & CLIP (zero-shot)   & touch-surface gating      & \Cref{app:p-clip} \\
Extraction  & GPT-5.5            & one call per video or window & \Cref{app:p-extract} \\
Grounding   & Qwen3.7-Plus      & first pass, no reasoning  & \Cref{app:p-ground} \\
Grounding   & Qwen3.7-Plus      & escalation, reasoning on  & \Cref{app:p-ground} \\
Grounding   & Qwen3.6-27B (local) & crop verification        & \Cref{app:p-ground} \\
\midrule
Training    & ---                & tool schema in context    & \Cref{app:actionspace} \\
\bottomrule
\end{tabular}
\end{table}

\subsection{Touch-surface Gating}
\label{app:p-clip}

Mobile recordings are screened with a zero-shot CLIP classifier using eight sampled frames, two positive captions, and four negative captions. A frame counts as mobile UI when a positive caption scores above all negatives. Videos are retained when at least four frames pass and the transcript contains at least two instructional phrases, detected using a fixed imperative-pattern list.

\lstinputlisting[caption={Positive and negative CLIP text prompts for mobile UI filtering.}]%
{sections/prompts/clip_gating.txt}

\subsection{Trajectory Extraction}
\label{app:p-extract}

The extraction model receives video metadata, transcript, and both frame streams in one call per video or temporal window. Desktop, web, and mobile prompts share task segmentation and action-extraction instructions but constrain outputs to their respective platforms. Additional rules discourage ambiguous target descriptions, incorrect drag endpoints, and unsupported scroll magnitudes.

\lstinputlisting[caption={Trajectory extraction prompt for desktop tutorial videos.}]%
{sections/prompts/extract_desktop.txt}

\lstinputlisting[caption={Trajectory extraction prompt for web tutorial videos.}]%
{sections/prompts/extract_web.txt}

\lstinputlisting[caption={Trajectory extraction prompt for mobile tutorial videos.}]%
{sections/prompts/extract_mobile.txt}

\subsection{Spatial Grounding}
\label{app:p-ground}

Grounding uses a high-recall first pass and a more conservative reasoning-enabled escalation pass (\Cref{sec:grounding}). The latter explicitly allows target absence rather than inferring coordinates from layout priors. A separate local model verifies uncertain predictions using a crop around the proposed target.

\lstinputlisting[caption={Initial spatial grounding prompt for point-coordinate prediction.}]%
{sections/prompts/ground_fast_point.txt}

\lstinputlisting[caption={Initial spatial grounding prompt for bounding-box prediction of text-entry targets.}]%
{sections/prompts/ground_fast_bbox.txt}

\lstinputlisting[caption={Reasoning-enabled spatial grounding prompt for point-coordinate verification.}]%
{sections/prompts/ground_esc_point.txt}

\lstinputlisting[caption={Reasoning-enabled spatial grounding prompt for bounding-box verification.}]%
{sections/prompts/ground_esc_bbox.txt}

\lstinputlisting[caption={Local crop-verification prompt for checking target presence.}]%
{sections/prompts/ground_verify.txt}

\subsection{Training-time System Prompts}

Training-time tool definitions are rendered from the action schema registry (\Cref{sec:actionspace}), keeping prompts consistent with normalized tool calls. Evaluation uses the same rendering, and each result records a SHA-256 prompt hash for reproducibility.

\subsection{Evaluation Prompts}
\label{app:p-eval}

The evaluation prompts are reproduced from the evaluation code (\Cref{sec:setup,sec:agentnet-heldout}). \Cref{tab:evalprompts} identifies the prompt used by each model and benchmark. Braced fields are filled per example, and \texttt{<image>} marks screenshot insertion.

\begin{table}[htbp]
\centering\small
\caption{Evaluation prompt used for each benchmark and model.}
\label{tab:evalprompts}
\begin{tabular}{lll}
\toprule
\textbf{Benchmark} & \textbf{Model} & \textbf{Prompt} \\
\midrule
OSWorld-G, ScreenSpot-Pro & Base   & Qwen3-VL cookbook computer use (\Cref{lst:eval-ground-base}) \\
OSWorld-G, ScreenSpot-Pro & T2T-8B & T2T desktop grounding (\Cref{lst:eval-ground-t2t}) \\
AndroidControl            & Base, T2T-8B & T2T mobile navigation (\Cref{lst:eval-ac}) \\
AgentNetBench             & Base   & AgentNetBench Qwen-VL agent (\Cref{lst:eval-anb-base}) \\
AgentNetBench             & T2T-8B & T2T desktop navigation (\Cref{lst:eval-anb-t2t}) \\
\bottomrule
\end{tabular}
\end{table}

\paragraph{Grounding.}
Base uses the official Qwen3-VL cookbook computer-use prompt with a $1000\times1000$ display convention; T2T-8B uses its training-time grounding template.

\lstinputlisting[caption={Evaluation prompt for the base model on GUI grounding benchmarks, adapted from the official Qwen3-VL computer-use cookbook.},label={lst:eval-ground-base}]%
{sections/prompts/eval_grounding_base.txt}

\lstinputlisting[caption={Evaluation prompt for T2T-8B on GUI grounding benchmarks, using the T2T desktop grounding template.},label={lst:eval-ground-t2t}]%
{sections/prompts/eval_grounding_t2t.txt}

\paragraph{AndroidControl.}
Both models use the T2T mobile navigation template, with an added \emph{Current step} field for the low-level instruction. The alternative template discussed in \Cref{sec:setup} additionally requires a \texttt{<think>} block before the \texttt{<action>} line.

\lstinputlisting[caption={Shared evaluation prompt for both models on AndroidControl, using the T2T mobile navigation template.},label={lst:eval-ac}]%
{sections/prompts/eval_androidcontrol.txt}

\paragraph{AgentNetBench.}
Base uses the benchmark's Qwen-VL agent prompt and \texttt{Step \{n\}: Thought:} prefill. T2T-8B uses its desktop navigation template. Both receive one current screenshot and textual action history.

\lstinputlisting[caption={Evaluation prompt for the base model on AgentNetBench, using the benchmark's official Qwen-VL agent template.},label={lst:eval-anb-base}]%
{sections/prompts/eval_agentnet_base.txt}

\lstinputlisting[caption={Evaluation prompt for T2T-8B on AgentNetBench, using the T2T desktop navigation template.},label={lst:eval-anb-t2t}]%
{sections/prompts/eval_agentnet_t2t.txt}

\FloatBarrier
\section{Scroll-Direction Statistics}
\label{app:scrolldir}

We report the scroll-direction distributions used in the in-domain experiment (\Cref{sec:indomain}). AndroidControl training scrolls are distributed $78.5/10.6/8.8/2.1\%$ (down/up/right/left); the additional training subset is rebalanced to $45/25/18/12\%$ without changing the proportion of scroll actions. In the test split, 1{,}211 scroll steps follow a $78.2/7.9/7.8/6.1\%$ distribution. Among steps correctly identified as scrolls, direction accuracy increases from 67.9\% to 87.8\% after in-domain fine-tuning. The predicted share of \texttt{down} rises from 65.8\% to 79.2\%, close to the test share of 78.2\%.